\documentclass[12pt]{article}

\usepackage{a4wide}
\usepackage{graphicx}
\usepackage{booktabs}
\usepackage{multirow}
\usepackage{array}
\usepackage{tabularx}
\usepackage{rotating}
\usepackage[dvipsnames]{xcolor}
\usepackage{amsmath, amssymb}
\usepackage{caption}
\usepackage{subcaption}
\usepackage[hidelinks]{hyperref}
\usepackage[numbers,sort&compress]{natbib}
\usepackage{url}
\usepackage{authblk}

\definecolor{greenlight}{RGB}{50,150,90}
\definecolor{lightblue}{RGB}{33, 140, 210}

\newcommand{\best}[1]{\textcolor{greenlight}{\textbf{#1}}}
\newcommand{\second}[1]{\textcolor{lightblue}{\textbf{#1}}}

\newcommand{\eg}{\textit{e.g.}}

\begin{document}

\title{OSMGraphCLIP: Learning Global Location Representations from OpenStreetMap Graphs}

\author[1]{Dimitrios Michail}
\author[2]{Eleni Saka}
\author[3]{Ioannis Giannopoulos}
\author[2,4]{Ioannis Papoutsis}

\affil[1]{Harokopio University of Athens, Athens, Greece.
  \texttt{michail@hua.gr}}
\affil[2]{National Technical University of Athens, Athens, Greece.
  \texttt{\{esaka,ipapoutsis\}@mail.ntua.gr}}
\affil[3]{Vienna University of Technology, Vienna, Austria.
  \texttt{igiannopoulos@geo.tuwien.ac.at}}
\affil[4]{National Observatory of Athens, Athens, Greece.}

\date{}

\maketitle

\begin{abstract}
We present OSMGraphCLIP, a CLIP-style geospatial representation model that learns global
location embeddings from freely available OpenStreetMap (OSM) data.
OSMGraphCLIP represents geographic environments as heterogeneous
graphs of typed OSM features, preserving the topological and semantic relationships among
roads, buildings, land-use regions, and points of interest. A multi-scale graph encoder
captures both fine-grained local structure and broader landscape composition, and supervises
a spherical-harmonics location encoder through a contrastive alignment objective. We
evaluate OSMGraphCLIP across a diverse suite of downstream geospatial regression and
classification tasks spanning climate, ecology, socioeconomic indicators, public health,
land cover, biodiversity, and wildfire forecasting, and show that structured OSM data
alone supports strong global location representations across domains.
OSMGraphCLIP matches or exceeds satellite-based baselines on the majority of benchmarks,
with the most pronounced advantage on socioeconomic and public-health tasks, where OSM's
explicit semantic annotation of the built environment encodes patterns of human activity
that satellite pixels can only capture indirectly. On ecological and environmental tasks,
the model remains closely competitive with imagery-based methods despite using no Earth
observation data. Qualitative analysis confirms that the learned embeddings organize
geographic space coherently, recovering biome boundaries, urban gradients, and
tropical--temperate distinctions from map topology alone.
\end{abstract}

\section{Introduction}
\label{sec:intro}

Understanding \emph{where} a location is, not just as a pair of coordinates, but in its full geographic, relational, environmental, landscape, and semantic context, is a foundational problem in geospatial machine
learning~\cite{mac2019presence,mai2023sphere2vec,klemmer2025satclip}. In geography, \emph{where} encodes more than absolute position, it captures spatial proximity, neighborhood structure, scale, accessibility, environmental conditions, land-use context, and the relationships between places. This perspective is closely aligned with Tobler's First Law of Geography, which states that nearby things tend to be more related than distant things.
A strong location representation should therefore capture both the intrinsic characteristics of a place and its spatial dependencies with surrounding locations. Such representations should transfer across a broad range of downstream tasks --- from predicting climate variables and ecological biomes
to inferring property values, species distributions, and population density
--- without requiring task-specific annotations.
Such representations underpin applications in ecology, public health, urban
planning, and earth observation, and have gained renewed interest as
geo-foundation models aim to match the transferability of large language and
vision models to the geospatial domain.

The dominant paradigm for learning globally transferable location
representations is \emph{contrastive alignment}: a location encoder that
maps geographic coordinates to a latent embedding is trained to produce
similar representations to those of a co-located context encoder.
GeoCLIP~\cite{geoclip2023} and SatCLIP~\cite{klemmer2025satclip} are standard instantiations
of this paradigm. GeoCLIP aligns a location encoder with a ground-level image encoder
using a CLIP loss~\cite{radford2021clip}, while SatCLIP extends this approach by aligning
geographic coordinates with co-located satellite imagery representations.
Both methods achieve strong results across diverse downstream tasks.

Satellite imagery is a natural context modality: it is globally available
and densely encodes land cover, vegetation, built-up patterns, and seasonal
phenology.
OpenStreetMap (OSM)~\cite{openstreetmap} offers a complementary perspective.
A prominent example of Volunteered Geographic Information (VGI), OSM is a freely
available, collaboratively curated global database contributed and maintained by a
large volunteer community.
Rather than capturing the spectral and physical characteristics of the Earth's surface,
OSM provides an explicit semantic description of geographic space: roads, buildings,
waterways, land-use polygons, and points of interest are annotated through a structured
\texttt{key=value} tagging vocabulary that captures their function and category ---
for example, \texttt{amenity:restaurant}, \texttt{landuse:residential}, or
\texttt{highway:motorway}.
Moreover, OSM features are inherently relational: roads intersect, buildings are
contained within land-use areas, rivers cross underneath bridges.
This native graph structure makes OSM particularly well suited for graph-based
representation learning, where heterogeneous nodes represent different types of
geographic entities and edges capture relations such as intersection, containment,
adjacency, proximity, and connectivity.

Despite these properties, OSM has been underutilized as a primary context
modality for global location representation learning.
Prior uses of OSM in geospatial models either rasterize vector data into
tile images~\cite{h3mosaic2025}, aggregate per-cell feature
counts~\cite{mora2025}, or focus on geo-entity retrieval rather than
globally transferable embeddings~\cite{bai2025geolink}.
Yet satellite imagery and OSM are naturally complementary modalities, and a key
open question is whether structured OSM data alone --- without any Earth observation
data --- is sufficient to learn strong global location representations.

We propose OSMGraphCLIP, which instead of an image encoder
utilizes a heterogeneous OSM graph encoder.
For each training location, we construct a heterogeneous graph of OSM
points, linestrings, and polygons within a chosen bounding box;
encode node semantics with pre-trained Sentence-BERT~\cite{reimers2019sbert} embeddings; and
derive topological edges from pairwise DE-9IM spatial relations~\cite{clementini1993small}.
Node representations are learned through heterogeneous graph attention
message passing across all cross-type spatial relations, yielding
relation-aware embeddings that capture both semantic and topological
structure. These representations are subsequently aggregated into a
fixed-dimensional graph-level embedding via hierarchical pooling and
attention-based readout mechanisms.
Finally, a Transformer band encoder aggregates coarser spatial context from
concentric regions at 2, 10, and 20km scales.
The fused embedding is then aligned with a spherical-harmonics
plus SIREN location encoder via a symmetric contrastive loss.

Our experiments show that structured OSM data alone is sufficient to
learn strong global location representations. Across 24 downstream tasks
spanning climate, ecology, land cover, biodiversity, socioeconomics, public
health, and wildfire forecasting, OSMGraphCLIP matches or exceeds
satellite-based baselines on the majority of benchmarks. The advantage is
most pronounced on socioeconomic and public-health prediction tasks, where
OSM's explicit semantic annotation of the built environment --- road
hierarchies, amenity categories, land-use designations --- encodes patterns
of human activity that satellite pixels can only capture indirectly. On
ecological and environmental tasks, OSMGraphCLIP remains closely competitive
with imagery-based methods despite using no Earth observation data,
demonstrating that the geographic structure of OSM features carries
substantial environmental signal. Qualitative analysis of the learned
embeddings confirms that the representations organize geographic space
coherently, recovering biome boundaries, urban gradients, and
tropical--temperate distinctions from map topology alone.

Our contributions are as follows.
\begin{itemize}
  \item We introduce OSMGraphCLIP, a CLIP-style geospatial
        representation model that learns globally transferable location
        embeddings without any satellite imagery, built on a novel architecture
        combining a heterogeneous OSM graph encoder with a
        multi-scale band encoder for broader spatial context.
  \item We construct a large-scale global pre-training corpus of
        approximately 200k geographically diverse locations and develop a
        scalable planetary extraction pipeline, including topology-aware
        heterogeneous graph construction via DE-9IM spatial relations and
        a density-stratified H3 sampling strategy for OSM-rich coverage.
    \item We demonstrate through extensive evaluation across 24
        downstream geospatial tasks that structured OSM graphs alone yield
        competitive global representations, with particular strength on
        semantics capture patterns of human activity that spectral imagery
        encodes only implicitly.
\end{itemize}

Code is publicly available at \url{https://github.com/d-michail/osmgraphclip};
pre-trained model checkpoints are released on Hugging Face.

\section{Related Work}
\label{sec:related}

\paragraph{Location encoding.}
Early work on coordinate-conditioned neural models focused on species
distribution modelling, learning spatial priors from geo-tagged biological
observations~\cite{mac2019presence}.
Space2Vec~\cite{mai2020space2vec} introduced multi-scale grid-cell
representations for spatial feature distributions, and
Sphere2Vec~\cite{mai2023sphere2vec} generalized this to the sphere,
proposing a family of multi-scale encodings for global geospatial
prediction.
Rußwurm et al.~\cite{russwurm2024spherical} further developed spherical
harmonic basis functions combined with SIREN activations as a stand-alone
location encoder, forming the backbone subsequently adopted by
SatCLIP~\cite{klemmer2025satclip} and inherited unchanged in our work.
SatCLIP introduced the contrastive alignment paradigm for global location
embeddings, pairing this location encoder with a ResNet satellite image
encoder and evaluating on ten downstream tasks spanning climate, ecology,
and socioeconomics.
Our work inherits the SatCLIP location encoder architecture and contrastive
training objective, but replaces satellite imagery with structured OSM
graphs as the context modality.

\paragraph{POI-based urban representations.}
Point-of-interest (POI) data provides a complementary, human-centred view
of geographic space: each POI conveys the semantic role of a location ---
education, commerce, healthcare --- through its category and name.
A large body of work has exploited spatial co-occurrence and graph-based
functional connectivity among POIs to produce semantic urban embeddings,
covering early embedding models through to graph neural network approaches.
The recent CaLLiPer framework~\cite{wang2025caliper,liu2025enriching} advances this line by
aligning POI text embeddings generated by large language models with
geographic coordinates, enabling richer semantic representations that
transfer to urban function mapping and mobility tasks.
While these POI-centric approaches demonstrate the value of semantic
annotation data, they remain confined to point geometries and do not
exploit the topological relationships --- roads crossing waterways, shops
lying within commercial zones, buildings sharing walls --- that OSM encodes
natively as a graph.
OSMGraphCLIP treats the full OSM feature set as a typed heterogeneous graph,
capturing relational topology alongside point-level semantics.

\paragraph{Multimodal contrastive alignment.}
To overcome the limitations of single-modality representations, recent work
aligns heterogeneous geospatial signals within a shared embedding space
using contrastive objectives inspired by CLIP~\cite{radford2021clip}.
GeoCLIP~\cite{geoclip2023} aligns GPS coordinates with satellite imagery
for worldwide geo-localization.
UrbanCLIP~\cite{urbanclip} employs language-generated captions for satellite
imagery to inject textual semantics into visual encoders.
MoRA~\cite{mora2025} generalizes contrastive alignment by coupling human
mobility traces with multimodal urban signals, exploiting mobility as a
structural backbone for scalable geospatial representation learning.
GAIR~\cite{gair2025} aligns satellite imagery, street-view imagery, and
geographic coordinates via contrastive learning, finding that combining
visual modalities yields stronger representations.
Population dynamics foundation models~\cite{agarwal2024geospatial} leverage
diverse multi-modal spatial signals for globally consistent demographic
inference.
Our work is distinct in using OSM exclusively \emph{as a graph} ---
preserving topology and cross-feature spatial relations --- rather than as
a source of feature counts, rasterized tiles, or textual tags.

\paragraph{OSM-based methods.}
Several methods exploit OSM data for region-level embedding via feature aggregation over H3 hexagons. Hex2vec~\cite{wozniak2021hex2vec} uses a skip-gram
objective over tag counts, Highway2vec~\cite{lesniara2022highway2vec} targets road
network characteristics, and GeoVeX~\cite{donghi2023geovex} scales to global
coverage with hexagonal convolutional autoencoders on tag-count histograms.
GeoLink~\cite{bai2025geolink} constructs heterogeneous OSM graphs with
point, line, and polygon nodes and trains them with contrastive and
reconstruction objectives for geo-entity linking and retrieval.
H3-MOSAIC~\cite{h3mosaic2025} combines OSM semantics with satellite imagery
by aggregating feature counts onto H3 grid cells, discarding intra-cell topology.
All these approaches reduce OSM features to per-cell counts or histograms,
losing cross-feature spatial relations.
Our heterogeneous graph attention encoder is inspired by GeoLink's graph
construction but is embedded in a global contrastive framework for
general-purpose location embeddings.

\begin{figure*}[t]
  \centering
  \includegraphics[width=\textwidth]{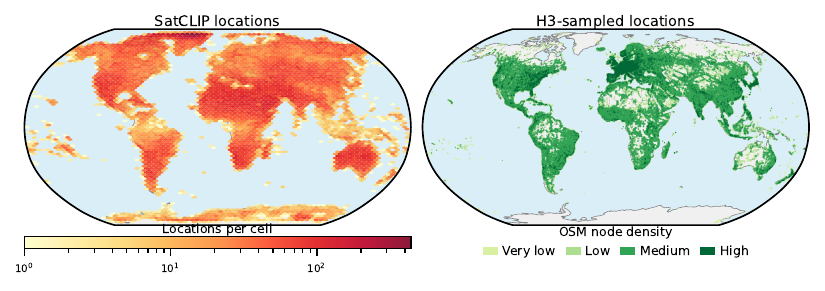}
  \captionsetup{skip=1pt}
  \caption{%
    Geographic distribution of the initial 200k candidate locations used to
    construct the OSMGraphCLIP corpus. The first 100k points are the globally
    distributed SatCLIP coordinates, and the remaining 100k are
    density-stratified H3-derived samples that emphasize OSM-rich regions
    while preserving global coverage. The final training set contains
    approximately 180k locations after preprocessing and quality filtering;
    the remaining approximately 20k locations are used for held-out
    validation.
  }
  \label{fig:location_dist}
\end{figure*}

\section{Methodology}
\label{sec:method}

\subsection{Location Selection}
\label{sec:locations}

Training OSMGraphCLIP requires a large, globally distributed set of geographic
coordinates for which OSM graphs can be constructed.
We assemble an initial corpus of 200k candidate locations from two
complementary sources,
each contributing 100k points.

The first 100k locations are the geographic coordinates used to train
SatCLIP~\cite{klemmer2025satclip}, drawn to achieve broad global coverage and
validated as a diverse benchmark corpus for contrastive location encoding.
Sharing this location set with SatCLIP enables direct like-for-like comparison:
both models observe the same query coordinates and differ only in the context
modality used to supervise the location encoder.

The second 100k locations are generated via a density-proportional sampling
strategy built around the H3 hierarchical hexagonal spatial
index~\cite{h3uber2018}.
We first identify globally data-bearing regions by estimating OSM feature
density over a coarse one-degree latitude/longitude grid, localizing the areas
that contain at least one annotated OSM feature.
Within each non-empty grid cell, we enumerate all H3 resolution-5 hexagonal
cells, which have an average diameter of approximately 22\,km and collectively
partition the globe into roughly two million non-overlapping hexagons.
The geographic center of each candidate cell is perturbed by a small random
jitter of $\pm 5$\,km to break the regular hexagonal lattice and increase
coordinate diversity.
Each candidate is then scored by the number of OSM features (points,
linestrings, and polygons) within a 500\,m radius, providing a spatially
localized measure of annotation density.
Candidates are stratified into four density tiers --- very low, low, medium,
and high --- and the target 100k locations are drawn by stratified sampling
with predetermined fractions that upweight informationally rich areas while
retaining global coverage.
A spatial cap of at most 10 selected locations per H3 resolution-3 cell
(approximately 130\,km in diameter) prevents geographic clustering and ensures
the selected set spans diverse landscapes, from dense urban cores to sparsely
annotated wilderness.

The two subsets are complementary by design: the SatCLIP coordinates provide
a well-validated, globally balanced sample anchored to an established
benchmark, while the H3-derived locations concentrate sampling on OSM-rich
regions --- cities, transport corridors, and agricultural landscapes --- while
still covering sparsely mapped environments via density-stratified selection.
The final training set used for contrastive learning contains approximately 180k locations.
The remaining approximately 20k locations from the initial 200k corpus are
used as a held-out validation set. See Figure~\ref{fig:location_dist}.

\subsection{Bounding Boxes and Graph Construction}
\label{sec:graph_construction}

For each training location $(\phi, \lambda)$, an axis-aligned bounding box
of half-width $b$ centred on the coordinate defines the spatial extent from
which OSM features are retrieved and assembled into a graph.
Constructing training graphs for this globally distributed corpus requires
managing geospatial data at planetary scale.
We ingested the full OpenStreetMap planet dump (on 24th March 2026) into a
local PostgreSQL/PostGIS instance using \texttt{osm2pgsql}, producing a
database of 847\,GB on disk.
The import uses the hstore extension to preserve all key-value tags and
creates four geometry tables---\texttt{planet\_osm\_point},
\texttt{planet\_osm\_line}, \texttt{planet\_osm\_polygon}, and
\texttt{planet\_osm\_roads}---each indexed
on a native Spherical Mercator (EPSG:3857) geometry column.
Load-time PostgreSQL configuration (WAL durability disabled, 40\,GB shared
buffer pool, 18 parallel import workers) reduces total planet ingest to
approximately 12--24 hours on commodity server hardware.

For each training location a bounding-box query is issued against all three
geometry tables.
Geometries are clipped to the query envelope via \texttt{ST\_ClipByBox2D}
before transfer, preventing full-resolution retrieval of large polygons such
as national parks or administrative boundaries.
Results are reprojected to WGS\,84 (EPSG:4326) with \texttt{ST\_Transform}
and serialized as compressed GeoJSON.
A per-table row cap of 50\,000 prevents pathological queries over the densest
urban cores: when the cap is reached the current location-scale pair is
flagged and the next coarser bounding-box level is substituted automatically.

The retrieved features are organized into a heterogeneous graph following GeoLink~\cite{bai2025geolink}, which represents the geographic
environment as a typed graph whose nodes correspond to geographic objects and
whose edges encode their spatial relationships.
Retrieved OSM features are partitioned into three node types: \emph{points}
(amenities, infrastructure, and points of interest), \emph{linestrings}
(roads, paths, and watercourses), and \emph{polygons} (buildings, parks, and
land-use zones).
Node features concatenate a semantic embedding with normalized geometric
descriptors.
For the semantic component, all available \texttt{key:value} tag pairs
associated with each feature are combined into a textual description and
encoded by a pre-trained Sentence-BERT model~\cite{reimers2019sbert}
into a fixed-dimensional dense vector. For example, a feature may be
represented by a sentence constructed from tags such as
\texttt{amenity:restaurant}, \texttt{cuisine:italian}, and
\texttt{outdoor\_seating:yes}. This differs from GeoLink, which encodes
each tag-value pair independently with a BERT model and aggregates the
resulting embeddings via a frequency-weighted mean. By jointly encoding
the full tag context with a sentence-level model, our approach captures
interactions between tags, reduces sensitivity to tag-frequency
imbalances, and generalizes more naturally to previously unseen tag
combinations.
The geometric component encodes the spatial position and shape of each
object within the bounding box: normalized coordinates for points, centroid
and endpoint positions for linestrings, and interior sample points for
polygons.

\begin{figure*}[t]
  \centering
  \includegraphics[width=\textwidth]{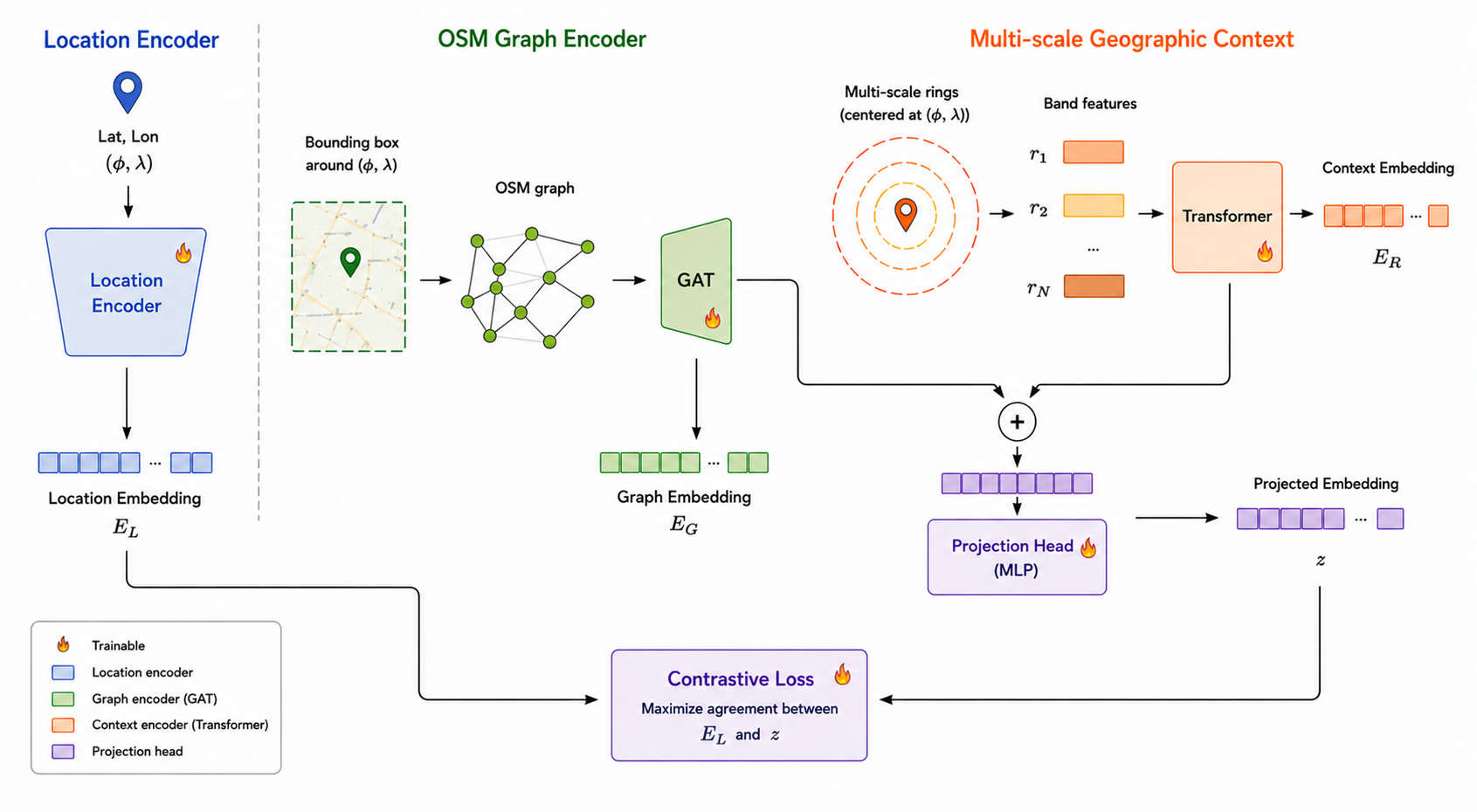}
  \captionsetup{skip=1pt}
  \caption{%
    OSMGraphCLIP overview. Given a geographic coordinate, a bounding box of OSM features is retrieved
    and encoded as a heterogeneous graph by a GAT-based encoder.
    Concentric radial bands at 2, 10, and 20\,km provide broader contextual
    signals via a Transformer band encoder.
    The fused embedding is contrastively aligned with a spherical-harmonics
    location encoder, producing location embeddings that transfer to diverse
    downstream tasks without any satellite imagery.
  }
  \label{fig:overview}
\end{figure*}

Edges encode pairwise topological structure using the DE-9IM spatial predicate
framework~\cite{de1998spatial}.
For each feature pair, predicates including \emph{touches},
\emph{overlaps}, \emph{covers}, \emph{covered-by}, \emph{crosses}, and
\emph{within} are evaluated according to geometry type; only non-disjoint
pairs yield a typed edge.
Topological connectivity is preferred over distance-based proximity as it
captures definitive spatial relationships that are invariant to scale
differences across geometry types.
For point-to-point pairs, where topological predicates are uninformative,
a Delaunay triangulation establishes neighborhood connectivity.
The resulting heterogeneous graph has three node types and nine directed
edge-relation types spanning all combinations of
$\{\text{point},\,\text{line},\,\text{polygon}\}^2$.

\begin{table}[th]
  \centering
  \caption{Components of the composite richness score $s$.
    Each component is independently normalized to $[0,1]$ before weighting.}
  \label{tab:richness}
  \begin{tabularx}{\textwidth}{@{} p{2.8cm} X c p{3.8cm} @{}}
    \toprule
    Component & Description & Weight & Normalization \\
    \midrule
    Tag entropy
      & Shannon entropy $H$ over (key, value) frequencies
      & 0.30 & $H / \log(50)$, cap 1 \\
    Category coverage
      & Fraction of 10 semantic categories present
      & 0.25 & $\in [0,1]$ \\
    Tag vocabulary
      & Number of distinct (key, value) pairs
      & 0.15 & $N_{\mathrm{kv}} / 30$, cap 1 \\
    Geometry entropy
      & Shannon entropy over point/line/polygon counts
      & 0.10 & $H / \log(3)$, cap 1 \\
    Spatial coverage
      & Fraction of $4{\times}4$ grid cells containing $\ge1$ feature
      & 0.10 & $\in [0,1]$ \\
    Semantic depth
      & Mean number of non-null tags per feature
      & 0.05 & depth$/8$, cap 1 \\
    IDF score
      & Mean IDF weight of present tag keys
      & 0.05 & IDF$/4$, cap 1 \\
    \midrule
    \textbf{Score} $s$ & Weighted sum & 1.00 & $[0,1]$ \\
    \bottomrule
  \end{tabularx}
\end{table}

\subsection{Multi-scale Spatial Encoding}
\label{sec:encoding_variants}
\label{sec:multiresolution}
\label{sec:band_encoder}

A central design question is how to determine the spatial scale at which to
construct the OSM graph for each training location and whether to incorporate
geographic context that extends beyond the local bounding box.
OSM annotation density varies enormously across the globe: a 250\,m box in a
dense urban center may contain hundreds of annotated features, whereas the
same box in a remote rural area may be nearly empty, making any single fixed
bounding-box size ill-suited for globally uniform representation learning.
At the same time, many geospatial properties --- large-scale land cover,
regional transport accessibility, proximity to major ecological features ---
are determined by context extending tens of kilometres beyond any single
bounding box.
We address these challenges with two complementary encoding strategies
corresponding to the \emph{base adaptive} and \emph{multiscale} model variants
evaluated in this paper.

\paragraph{Adaptive resolution sampling.}
The base adaptive variant constructs one OSM graph per training location by selecting
the most semantically informative bounding-box scale from a discrete set of
five candidate half-widths: L0 (200\,m), L1 (500\,m), L2 (2\,000\,m),
L3 (5\,000\,m), and L4 (20\,000\,m), probed from finest to coarsest.
For each (location, scale) pair, a composite \emph{richness score}
$s \in [0,1]$ is computed as a weighted aggregation of seven complementary indicators of
semantic content (Table~\ref{tab:richness}).
The indicator weights were calibrated in pilot preprocessing runs with the
explicit objective of selecting graphs that are informative but not
pathologically sparse, targeting a typical graph size of roughly 10--20
nodes per selected location.
The calibration criterion was based on graph-construction statistics only
(node-count distribution and empty/sparse-graph rates), and did \emph{not}
optimize downstream benchmark performance.
The scale achieving the highest score above a minimum threshold
$s_{\min} = 0.2$ is retained; if no scale meets this threshold, the bounding
box is expanded multiplicatively by a factor of up to $\alpha = 3.0$ beyond
the coarsest scale.
If any scale is reached with zero features retrieved, the strategy escalates
automatically to the next coarser profile, ensuring that data-sparse
locations --- deserts, tundra, or open water --- contribute a training sample
rather than being silently discarded.

On the final processed corpus used for graph construction
($n=198{,}323$ locations), this policy produces no empty graphs, with 89.6\%
of selected graphs containing at least 10 nodes, median node count 14
($p_{25}=11$, $p_{75}=26$).
The selected-scale distribution is broad rather than collapsing to a single
resolution: L0 (200\,m) 32.5\%, L1 (500\,m) 15.4\%, L2 (2\,000\,m) 23.2\%,
L3 (5\,000\,m) 13.6\%, L4 (20\,000\,m) 14.7\%, and L5 (adaptive expansion)
0.7\%.

\paragraph{Multi-scale band encoder.}
The multiscale variant uses a different strategy. Instead of trying to figure
out the correct bounding-box scale, it fixes the fine-grain graph to a bounding-box
half-width of $b = 1$\,km and supplements it with a Transformer-based
\emph{band encoder} that aggregates coarser context from three concentric
radial bands centred on the query location, at radii $r_1 = 2$\,km,
$r_2 = 10$\,km, and $r_3 = 20$\,km.
Rather than selecting a single bounding-box scale, this variant explicitly
represents context at multiple spatial resolutions simultaneously: the local
graph captures neighborhood topology and semantic detail, while the
concentric bands provide progressively coarser summaries of the surrounding
landscape.
The pipeline retrieves all features within the outermost radius in a single
database query and partitions the result in-process by centroid distance,
reducing total database round-trips to one per location.

For each band radius $r_b$, all OSM features within a disk of radius $r_b$
are queried and three levels of spatial statistics are computed:
\begin{itemize}
  \item \textbf{Global band features} (47-dimensional): aggregate counts by
        semantic category; feature density and spatial dispersion; tag
        vocabulary entropy; and the composite richness score of the full disk.
  \item \textbf{Sub-bin features} (16-dimensional per sub-bin): the same
        statistics computed independently for the inner ring (the annulus
        between $r_{b-1}$ and $r_b$) and the outer ring, capturing radial
        gradients in land use and feature density.
  \item \textbf{Sector features} (11-dimensional per sector): per-cardinal-%
        direction aggregates for the four quadrants (N, E, S, W), capturing
        directional asymmetries (\eg a location near a coast or urban edge).
\end{itemize}
In addition, a distance-weighted SBERT semantic embedding is computed for
each partition: each feature's tag embedding is weighted by
$\exp(-d/r_b)$, where $d$ is its distance from the query location, and
the weighted mean serves as the semantic summary.

\subsection{Model Architecture}
\label{sec:model}

Figure~\ref{fig:overview} provides an overview of the OSMGraphCLIP architecture.
All three encoder components share an embedding dimension $d = 256$.

\paragraph{Graph encoder.}
A single heterogeneous graph attention layer~\cite{velickovic2018graph} with
independent weight matrices per directed edge-relation type encodes the OSM
node features, producing updated node embeddings
$\mathbf{H}^t \in \mathbb{R}^{n_t \times 256}$ for each type
$t \in \{\text{point}, \text{line}, \text{polygon}\}$.
Set2Set~\cite{set2set2016} pooling ($T=5$ steps) followed by a per-type
linear projection yields a 256-dimensional aggregated representation
$\mathbf{a}^t$ per type; the three per-type vectors are combined by learned
attention to form the graph embedding:
\begin{equation}
  \mathbf{g} = \mathbf{W}_{\text{proj}}
               \sum_{t} \frac{\exp(\mathbf{w}^\top \mathbf{a}^t)}
                             {\sum_{t'}\exp(\mathbf{w}^\top \mathbf{a}^{t'})}
               \, \mathbf{a}^t \in \mathbb{R}^d,
\end{equation}
allowing the model to down-weight geometry types that are uninformative for
a given location.

\paragraph{Band attention encoder (multiscale variant).}
The spatial statistics and SBERT semantic embeddings for each band partition
are concatenated and linearly projected to $d_{\text{model}} = 256$:
\begin{align}
  \mathbf{t}^{\text{global}}_b
    &= \mathbf{W}_g\,[\mathbf{f}^{\text{global}}_b \,\|\, \mathbf{e}^{\text{global}}_b],
  \\
  \mathbf{t}^{\text{sub}}_{b,k}
    &= \mathbf{W}_s\,[\mathbf{f}^{\text{sub}}_{b,k} \,\|\, \mathbf{e}^{\text{sub}}_{b,k}],
    \quad k \in \{1,2\},
  \\
  \mathbf{t}^{\text{sec}}_{b,\ell}
    &= \mathbf{W}_q\,[\mathbf{f}^{\text{sec}}_{b,\ell} \,\|\, \mathbf{e}^{\text{sec}}_{b,\ell}],
    \quad \ell \in \{N,E,S,W\}.
\end{align}
These 22 tokens --- a learnable CLS token, $n_b = 3$ global-band tokens,
$2n_b = 6$ sub-bin tokens, and $4n_b = 12$ sector tokens, each with a
learned positional embedding --- are processed by a two-layer Transformer
encoder ($h=4$ heads, feed-forward dimension $1{,}024$).
The CLS output is projected to $\mathbb{R}^d$:
\begin{equation}
  \mathbf{b} = \mathbf{W}_{\text{out}}\,
               \text{TransformerEncoder}(\mathbf{t}_\text{CLS}) \in \mathbb{R}^d.
\end{equation}
The band and graph embeddings are fused by concatenation and projection:
\begin{equation}
  \mathbf{z}_{\text{fused}} = \mathbf{W}_{\text{fuse}}\,[\mathbf{g} \,\|\, \mathbf{b}]
  \in \mathbb{R}^d.
\end{equation}
In the base variant, $\mathbf{g}$ itself serves as the context embedding
without the band encoder.

\paragraph{Location encoder.}
We adopt the SatCLIP location encoder without
modification~\cite{klemmer2025satclip}.
Geographic coordinates are lifted to the unit sphere and spherical harmonic
basis functions are evaluated up to Legendre polynomial degree $L$, yielding
a $2(L+1)^2$-dimensional positional encoding.
We experiment with $L=10$ (242 dimensions, denoted L10) and $L=40$
(3362 dimensions, denoted L40).
A SIREN~\cite{sitzmann2020siren} with two hidden layers of width 512 maps
this encoding to $\mathbb{R}^d$.

\section{Experiments}
\label{sec:experiments}

We train four OSMGraphCLIP models defined by two architectural choices
and two spherical-harmonic resolutions $L \in \{10, 40\}$:
\begin{itemize}
  \item \textbf{OSMGraphCLIP-A-L\{10,40\}}: the base \emph{adaptive} variant,
        which uses adaptive resolution graph encoding,
  \item \textbf{OSMGraphCLIP-MS-L\{10,40\}}: the \emph{multiscale} variant,
        which combines the fixed-scale graph encoder with the band attention
        encoder.
\end{itemize}
We use the shorthand \textbf{A-L10}, \textbf{A-L40}, \textbf{MS-L10}, and
\textbf{MS-L40} when referring to these four models in tables and figures.

All models share the symmetric CLIP-style contrastive objective adopted
from SatCLIP~\cite{radford2021clip,klemmer2025satclip}.
For a batch of $N$ (context, coordinate) pairs, $L_2$-normalized context
embeddings $\{\mathbf{z}_i\}$ and location embeddings $\{\mathbf{c}_i\}$
form a cosine-similarity logit matrix
$L_{ij} = \tau^{-1}\mathbf{z}_i^\top\mathbf{c}_j$ with learnable
temperature $\tau$.
Training minimises the mean of two symmetric cross-entropy terms:
\begin{equation}
  \mathcal{L} = \tfrac{1}{2}\bigl[\mathcal{L}_{\text{ctx}} + \mathcal{L}_{\text{loc}}\bigr], \quad
  \mathcal{L}_{\text{ctx}} = -\tfrac{1}{N}\sum_{i=1}^N \log
  \frac{\exp(L_{ii})}{\sum_{j=1}^N \exp(L_{ij})},
  \label{eq:loss}
\end{equation}
with $\tau$ initialized at $0.07$ (log-scale surrogate clamped to
$[0,\log(100)]$).
All models are optimized with AdamW~\cite{loshchilov2017adamw} (learning
rate $10^{-4}$, weight decay $0.01$) with a maximum budget of 3000 epochs.
For each run, we select the checkpoint with the lowest contrastive loss on
a held-out validation set of approximately 20k locations, early stopping
by model selection at convergence.
The embedding dimension is $d = 256$, shared between the graph, band, and
location encoders.
Biases, normalization parameters, and the temperature $\tau$ are excluded
from weight decay.
All four models use SBERT-384 node embeddings.

Table~\ref{tab:msl40_training} summarises the full training trajectory of
the \textbf{MS-L40} model, which is our primary configuration.
The run used approximately 180k training locations with batch size 8{,}192
($\approx$22 optimization steps per epoch), and converged at epoch 1{,}746
(validation loss 4.56), while training was allowed to continue until epoch
2{,}017 without further improvement.
The last restarts repeatedly resumed from the best checkpoint and confirmed
that performance had saturated.
The other three variants (A-L10, A-L40, MS-L10) were trained with the same
protocol and checkpoint-selection criterion.

\begin{table}[t]
  \centering
  \caption{Training summary for OSMGraphCLIP MS-L40.}
  \label{tab:msl40_training}
  \begin{tabularx}{\columnwidth}{@{}p{0.43\columnwidth}X@{}}
    \toprule
    Statistic & Value \\
    \midrule
    Trainable parameters & 7.9M \\
    Hardware / precision & 1$\times$ NVIDIA H200,\newline bfloat16 mixed precision \\
    Training locations & $\approx$180k \\
    Validation locations & $\approx$20k \\
    Batch size / steps per epoch & 8{,}192 / $\approx$22 \\
    Max allowed epochs & 3{,}000 \\
    Executed epochs & 2{,}017 \\
    Best checkpoint (val loss) & epoch 1{,}746 (4.56) \\
    Total optimization steps & $\approx$44{,}374 \\
    Time per epoch & $\approx$68 s (50 s train + 18 s val) \\
    Active GPU compute & 70 GPU-hours\\
    Wall-clock span & $\approx$77.5 h\\
    \bottomrule
  \end{tabularx}
\end{table}

\subsection{Downstream Evaluation Protocol}
\label{sec:evaluation}

We compare our models against five baselines, all evaluated under a unified protocol.
GeoCLIP~\cite{geoclip2023} aligns ground-level images with GPS coordinates via a CLIP-inspired contrastive objective.
AlphaEarth~\cite{brown2025alphaearth} generates a unified 64-dimensional representation for every 10\,m² land-surface patch by assimilating multi-modal Earth observation data (Sentinel-2, Landsat, Sentinel-1 SAR, GEDI LiDAR, and geotagged text) via a multi-objective reconstruction objective; due to its land-only coverage it is excluded from tasks requiring ocean-location embeddings (Countries, Biomes, Ecoregions).
SatCLIP-L10 and SatCLIP-L40~\cite{klemmer2025satclip} are location encoders trained by contrastively aligning GPS coordinates with Sentinel-2 imagery using a spherical-harmonics encoder at Legendre degrees $L{=}10$ and $L{=}40$, evaluated here with a ResNet-50 backbone.
GT-Loc~\cite{shatwell2025gt} jointly predicts geo-location and capture timestamp from images by aligning image, time, and location embeddings in a shared space via a cyclical metric-learning objective over a toroidal surface.
Copernicus-FM~\cite{wang2025unifiedcopernicusfoundationmodel} is an EO foundation model pre-trained on 18.7M aligned images from all major Copernicus Sentinel missions using dynamic hypernetworks for flexible multi-sensor and metadata encoding.

We consider 24 downstream geospatial prediction tasks: 9 SatCLIP benchmarks
following Klemmer et al.~\cite{klemmer2025satclip}, 3 additional geospatial
benchmarks (SatBird, reBEN, and wildfire forecasting), and 12 CDC PLACES
health regression tasks~\cite{cdcplaces2023} following Agarwal et al.~\cite{agarwal2024geospatial}.
For each task and encoder, we use a two-layer MLP head (hidden size 128, dropout 0.5). In all cases,
inputs to the downstream MLP are raw latitude--longitude embeddings with no task-specific features.
Exceptions are iNaturalist, where location embeddings are concatenated with pretrained InceptionV3
image features, and wildfire forecasting, where a cyclic day-of-year encoding is added to capture
seasonality (see Appendix~\ref{app:Evaluation}).

\paragraph{Regression tasks ($R^2$).}
\textit{Air temperature}: annual mean near-surface air temperature from the
global station-and-satellite dataset of Hooker et al.~\cite{hooker2018airtemp}.
\textit{Elevation}: terrain elevation from the dataset compiled by Rolf et
al.~\cite{rolf2021generalizable} as part of the SustainBench benchmark.
\textit{Median income}: median household income for census tracts in the
contiguous United States, from Jia and Benson~\cite{jia2020residual}.
\textit{California housing}: residential property prices from the spatial
econometrics benchmark of Pace and Barry~\cite{pace2003semiparametric}.
\textit{Population density}: logged population density from the SustainBench
compilation of Rolf et al.~\cite{rolf2021generalizable}.

\paragraph{SatBird (top-$k$ accuracy)}
Bird species encounter-rate prediction from the SatBird benchmark~\cite{satbird_NEURIPS2023}
(USA-summer subset), evaluated using the official top-$k$ retrieval accuracy protocol.
Unlike $R^2$, which is poorly suited to sparse multi-species targets where near-zero predictions
can appear accurate for rare classes, top-$k$ accuracy directly measures whether the model
recovers the most likely species at a location, making it more appropriate for species
distribution evaluation.

\paragraph{Health tasks ($R^2$).}
Twelve public-health outcome measures sourced from the CDC PLACES 2023
release~\cite{cdcplaces2023}, which provides model-based small-area prevalence
estimates at the ZIP Code Tabulation Area (ZCTA) level for the contiguous
United States.
Following the evaluation approach of Agarwal et al.~\cite{agarwal2024geospatial},
who benchmark a population dynamics foundation model on 21 health outcomes, we
use a subset of twelve measures as regression targets:
\textit{Physical health not good},
\textit{Diabetes},
\textit{Chronic obstructive pulmonary disease (COPD)},
\textit{Cancer} (excluding skin cancer),
\textit{Coronary heart disease},
\textit{Mental health not good},
\textit{Received annual checkup},
\textit{Sleep less than 7 hours},
\textit{Asthma},
\textit{Obesity},
\textit{Smoking} (current smokers), and
\textit{High cholesterol}.
All values are age-adjusted prevalence rates (percentage of adults). As these measures are restricted to the contiguous United States, this task group evaluates the encoders' ability to capture fine-grained socioeconomic and health-related geographic variation within a single country.

\paragraph{Classification tasks (accuracy).}
\textit{Country}: country-of-origin classification derived from coordinate
metadata, as introduced by Klemmer et al.~\cite{klemmer2025satclip}.
\textit{Biome} and \textit{Ecoregion}: biome- and ecoregion-level land
classifications derived from the global map of Dinerstein et
al.~\cite{dinerstein2017ecoregion}, which partitions the terrestrial realm
into 14 biomes and 846 ecoregions based on climate and biogeographic criteria.
\textit{reBEN}: multi-label land-cover classification
using the refined BigEarthNet benchmark~\cite{clasen2024reben},
which assigns Sentinel-2 patches to 19 land-cover categories.
We report micro-F1 following the standard benchmark protocol,
as patches may contain multiple labels and class frequencies
are highly imbalanced. \textit{iNaturalist}: species classification on a geographically stratified
subset of iNaturalist observations~\cite{horn2018inaturalist}.
Following Klemmer et al.~\cite{klemmer2025satclip} and the original geo-prior
evaluation protocol of Mac Aodha et al.~\cite{mac2019presence}, the location
embedding is concatenated with image features extracted from a pre-trained
InceptionV3 model before the MLP head is trained; this is the only task for which image-derived features are used.

\paragraph{Wildfire forecasting
(AUPRC/average precision)}:
Binary prediction of wildfire occurrence following
Mesogeos~\cite{kondylatos2023mesogeos}.
Because wildfire events are rare, overall accuracy is
uninformative under severe class imbalance.
We therefore report AUPRC, which evaluates performance across precision--recall
trade-offs.

\begin{table*}[!ht]
  \centering
  \caption{%
    Downstream task performance ($R^2$ for regression, incl.\ 12
    public-health outcomes; accuracy for classification),
    mean $\pm$ std over ten random seeds.
    All baselines are locally re-evaluated under the same MLP-head
    protocol.
    Type codes: S = Socioeconomic, E = Environment, H = Health.
    \best{Bold green} = best per row; \second{bold light blue} = second best.
  }
  \label{tab:main_results}
  \normalsize
  \setlength{\tabcolsep}{2.5pt}
  \renewcommand{\arraystretch}{1.1}
  \resizebox{\textwidth}{!}{%
  \begin{tabular}{@{}p{2.2cm} c cccccc cccc@{}}
    \toprule
    & & \multicolumn{6}{c}{Baselines} &
      \multicolumn{4}{c}{OSMGraphCLIP (ours)} \\
    \cmidrule(lr){3-8}\cmidrule(lr){9-12}
    \multirow{2}{*}{Task}
      & \multirow{2}{*}{Type}
      & GeoCLIP
      & AlphaEarth
      & SatCLIP-L10
      & SatCLIP-L40
      & GT-Loc
      & Copernicus
      & A-L10
      & A-L40
      & MS-L10
      & MS-L40 \\
    &
      & \cite{geoclip2023}
      & \cite{brown2025alphaearth}
      & ~\cite{klemmer2025satclip}
      & ~\cite{klemmer2025satclip}
      & \cite{shatwell2025gt}
      & FM~\cite{wang2025unifiedcopernicusfoundationmodel}
      & (adaptive)
      & (adaptive)
      & (multiscale)
      & (multiscale) \\
    \midrule
    \multicolumn{12}{l}{\textit{Regression ($R^2$; top-$k$ for SatBird)}} \\[2pt]
    Air temp
      & E
      & $0.942{\pm}.007$
      & \best{0.981}${\pm}.000$
      & $0.939{\pm}.001$
      & $0.938{\pm}.002$
      & $0.942{\pm}.001$
      & $0.906{\pm}.001$
      & \second{0.956}${\pm}.001$
      & $0.860{\pm}.008$
      & $0.955{\pm}.002$
      & $0.885{\pm}.004$ \\
    Median income
      & S
      & $0.450{\pm}.004$
      & $0.281{\pm}.006$
      & $0.500{\pm}.007$
      & $0.502{\pm}.006$
      & $0.458{\pm}.006$
      & $0.328{\pm}.010$
      & $0.455{\pm}.006$
      & \best{0.524}${\pm}.006$
      & $0.437{\pm}.006$
      & \second{0.519}${\pm}.010$ \\
    Cali housing
      & S
      & \second{0.773}${\pm}.003$
      & $0.548{\pm}.005$
      & $0.634{\pm}.005$
      & $0.633{\pm}.003$
      & \best{0.782}${\pm}.002$
      & $0.433{\pm}.001$
      & $0.439{\pm}.012$
      & $0.635{\pm}.006$
      & $0.504{\pm}.030$
      & $0.640{\pm}.005$ \\
    Elevation
      & E
      & $0.823{\pm}.003$
      & \best{0.978}${\pm}.000$
      & $0.897{\pm}.001$
      & $0.898{\pm}.001$
      & $0.838{\pm}.001$
      & \second{0.940}${\pm}.000$
      & $0.870{\pm}.001$
      & $0.887{\pm}.001$
      & $0.871{\pm}.002$
      & $0.880{\pm}.001$ \\
    Population
      & S
      & $0.781{\pm}.001$
      & $0.801{\pm}.000$
      & \best{0.821}${\pm}.001$
      & \second{0.821}${\pm}.001$
      & $0.785{\pm}.001$
      & $0.769{\pm}.002$
      & $0.812{\pm}.001$
      & $0.813{\pm}.002$
      & $0.812{\pm}.001$
      & $0.814{\pm}.001$ \\
    SatBird (top-$k$)
      & E
      & $0.632{\pm}.001$
      & \best{0.675}${\pm}.001$
      & $0.596{\pm}.001$
      & $0.607{\pm}.001$
      & \second{0.633}${\pm}.001$
      & $0.595{\pm}.000$
      & $0.543{\pm}.001$
      & $0.558{\pm}.001$
      & $0.551{\pm}.002$
      & $0.558{\pm}.001$ \\
    Phys.\ health
      & H
      & $0.597{\pm}.003$
      & $0.473{\pm}.005$
      & $0.530{\pm}.003$
      & $0.595{\pm}.003$
      & $0.610{\pm}.004$
      & $0.565{\pm}.006$
      & $0.545{\pm}.006$
      & \second{0.611}${\pm}.004$
      & $0.537{\pm}.003$
      & \best{0.620}${\pm}.004$ \\
    Diabetes
      & H
      & \second{0.480}${\pm}.005$
      & $0.354{\pm}.004$
      & $0.392{\pm}.005$
      & $0.466{\pm}.004$
      & $0.477{\pm}.003$
      & $0.422{\pm}.010$
      & $0.407{\pm}.004$
      & $0.479{\pm}.004$
      & $0.400{\pm}.007$
      & \best{0.487}${\pm}.004$ \\
    COPD
      & H
      & $0.612{\pm}.005$
      & $0.490{\pm}.003$
      & $0.571{\pm}.004$
      & $0.631{\pm}.004$
      & $0.625{\pm}.005$
      & $0.601{\pm}.007$
      & $0.585{\pm}.005$
      & \second{0.652}${\pm}.002$
      & $0.577{\pm}.005$
      & \best{0.655}${\pm}.003$ \\
    Cancer
      & H
      & \second{0.354}${\pm}.004$
      & $0.275{\pm}.003$
      & $0.254{\pm}.006$
      & $0.329{\pm}.006$
      & \best{0.360}${\pm}.007$
      & $0.298{\pm}.006$
      & $0.271{\pm}.008$
      & $0.342{\pm}.007$
      & $0.265{\pm}.006$
      & $0.351{\pm}.006$ \\
    Coronary HD
      & H
      & $0.473{\pm}.006$
      & $0.346{\pm}.003$
      & $0.410{\pm}.006$
      & $0.489{\pm}.003$
      & $0.475{\pm}.008$
      & $0.439{\pm}.007$
      & $0.434{\pm}.006$
      & \second{0.509}${\pm}.005$
      & $0.420{\pm}.005$
      & \best{0.513}${\pm}.007$ \\
    Ment.\ health
      & H
      & \second{0.525}${\pm}.005$
      & $0.394{\pm}.005$
      & $0.418{\pm}.006$
      & $0.465{\pm}.005$
      & \best{0.548}${\pm}.005$
      & $0.445{\pm}.005$
      & $0.428{\pm}.004$
      & $0.478{\pm}.004$
      & $0.424{\pm}.004$
      & $0.480{\pm}.007$ \\
    Ann.\ checkup
      & H
      & $0.766{\pm}.003$
      & $0.683{\pm}.004$
      & $0.749{\pm}.002$
      & $0.784{\pm}.003$
      & $0.768{\pm}.002$
      & $0.761{\pm}.002$
      & $0.757{\pm}.003$
      & \second{0.792}${\pm}.003$
      & $0.754{\pm}.003$
      & \best{0.796}${\pm}.002$ \\
    Sleep $<$7h
      & H
      & $0.639{\pm}.004$
      & $0.495{\pm}.004$
      & $0.552{\pm}.005$
      & $0.619{\pm}.005$
      & \best{0.659}${\pm}.002$
      & $0.592{\pm}.006$
      & $0.563{\pm}.004$
      & $0.637{\pm}.005$
      & $0.558{\pm}.004$
      & \second{0.650}${\pm}.005$ \\
    Asthma
      & H
      & \second{0.574}${\pm}.006$
      & $0.407{\pm}.007$
      & $0.542{\pm}.006$
      & $0.542{\pm}.006$
      & \best{0.601}${\pm}.006$
      & $0.513{\pm}.008$
      & $0.496{\pm}.004$
      & $0.552{\pm}.005$
      & $0.490{\pm}.005$
      & $0.559{\pm}.004$ \\
    Obesity
      & H
      & $0.618{\pm}.003$
      & $0.481{\pm}.003$
      & $0.616{\pm}.003$
      & $0.616{\pm}.003$
      & $0.633{\pm}.002$
      & $0.558{\pm}.004$
      & $0.564{\pm}.002$
      & \second{0.638}${\pm}.002$
      & $0.555{\pm}.003$
      & \best{0.642}${\pm}.004$ \\
    Smoking
      & H
      & $0.628{\pm}.003$
      & $0.503{\pm}.006$
      & $0.624{\pm}.004$
      & $0.623{\pm}.004$
      & \second{0.646}${\pm}.003$
      & $0.596{\pm}.007$
      & $0.574{\pm}.003$
      & $0.643{\pm}.003$
      & $0.572{\pm}.003$
      & \best{0.648}${\pm}.005$ \\
    High cholesterol
      & H
      & \second{0.496}${\pm}.005$
      & $0.349{\pm}.010$
      & $0.432{\pm}.004$
      & $0.432{\pm}.004$
      & \best{0.524}${\pm}.005$
      & $0.409{\pm}.006$
      & $0.407{\pm}.003$
      & $0.437{\pm}.004$
      & $0.405{\pm}.004$
      & $0.437{\pm}.007$ \\
    \midrule
    \multicolumn{12}{l}{\textit{Classification (accuracy; \ aver. precision for Wildfire)}} \\[2pt]
    Country
      & S
      & $0.899{\pm}.002$
      & N/A
      & \best{0.954}${\pm}.000$
      & \second{0.954}${\pm}.000$
      & $0.925{\pm}.001$
      & $0.835{\pm}.001$
      & $0.941{\pm}.001$
      & $0.947{\pm}.000$
      & $0.947{\pm}.000$
      & $0.950{\pm}.000$ \\
    iNaturalist
      & E
      & $0.448{\pm}.013$
      & $0.491{\pm}.009$
      & $0.563{\pm}.002$
      & \best{0.564}${\pm}.003$
      & $0.492{\pm}.012$
      & $0.403{\pm}.019$
      & $0.559{\pm}.003$
      & $0.546{\pm}.005$
      & \second{0.564}${\pm}.004$
      & $0.550{\pm}.003$ \\
    Biome
      & E
      & $0.896{\pm}.000$
      & N/A
      & \best{0.941}${\pm}.000$
      & \second{0.941}${\pm}.000$
      & $0.914{\pm}.001$
      & $0.856{\pm}.000$
      & $0.916{\pm}.000$
      & $0.906{\pm}.000$
      & $0.926{\pm}.000$
      & $0.915{\pm}.001$ \\
    Ecoregion
      & E
      & $0.822{\pm}.001$
      & N/A
      & \best{0.914}${\pm}.001$
      & \second{0.914}${\pm}.000$
      & $0.859{\pm}.001$
      & $0.775{\pm}.004$
      & $0.891{\pm}.001$
      & $0.896{\pm}.001$
      & $0.896{\pm}.001$
      & $0.890{\pm}.000$ \\
    reBEN
      & E
      & $0.559{\pm}.002$
      & $0.427{\pm}.004$
      & \second{0.573}${\pm}.006$
      & $0.572{\pm}.005$
      & $0.563{\pm}.003$
      & $0.573{\pm}.003$
      & $0.549{\pm}.005$
      & $0.569{\pm}.006$
      & $0.546{\pm}.004$
      & \best{0.574}${\pm}.003$ \\
    Wildfire (Avg.\ P)
      & E
      & \second{0.798}${\pm}.002$
      & \best{0.804}${\pm}.005$
      & $0.794{\pm}.003$
      & $0.790{\pm}.004$
      & $0.795{\pm}.005$
      & $0.749{\pm}.001$
      & $0.769{\pm}.003$
      & $0.784{\pm}.004$
      & $0.774{\pm}.004$
      & $0.794{\pm}.002$ \\
      \midrule
      1st/2nd best
      & --
      &\best{0} / \second{7}
      & \best{4} / \second{0}
      & \best{4} / \second{1}
      & \best{1} / \second{4}
      & \best{6} / \second{2}
      & \best{0} / \second{1}
      & \best{0} / \second{1}
      & \best{1} / \second{5}
      & \best{0} / \second{1}
      & \best{8} / \second{2} \\
    \bottomrule
  \end{tabular}%
}
\end{table*}

\subsection{Main Results}
\label{sec:main_results}

Table~\ref{tab:main_results} compares our best models against the baselines.
All OSMGraphCLIP variants are trained on the final approximately 180k-location set
(sampled from the initial 200k candidate corpus); the \emph{adaptive} variants use the adaptive single-scale
architecture while the \emph{multiscale} (MS) variants fix the resolution of
the bounding box for graph construction and add the band attention encoder.

OSMGraphCLIP-MS-L40 delivers the strongest overall performance, ranking first or second on 10 of 24 benchmark
entries — more than any other individual model — despite relying exclusively on OpenStreetMap data and excluding
satellite or Earth observation inputs. Performance gains are particularly pronounced for public-health and
socioeconomic tasks: MS-L40 achieves the best result on 7 of the 12 CDC PLACES outcomes and remains highly
competitive on core regression benchmarks, while A-L40 achieves the strongest performance on median income ($R^2 = 0.524$).
More broadly, OSMGraphCLIP consistently outperforms GeoCLIP, SatCLIP, and GT-Loc on tasks where characteristics of the
built environment are strongly predictive, suggesting that OSM's explicit semantic representation of roads, amenities, and
land use effectively captures socioeconomic structure.

OSMGraphCLIP also performs competitively on several geographic and environmental tasks. A-L10 achieves the strongest
performance among coordinate-based approaches on air temperature ($R^2 = 0.956$), trailing only AlphaEarth, which
incorporates multimodal Earth observation data. On country classification, MS-L40 reaches 0.950 accuracy, closely
matching SatCLIP-L40 (0.954), while remaining competitive on biome and ecoregion prediction
(MS-L10: 0.926 and 0.896, respectively). Performance on reBEN is broadly comparable across methods, with MS-L40
obtaining the highest micro-F1 score (0.574).

In contrast, OSMGraphCLIP underperforms imagery-based approaches on ecological and habitat-sensitive benchmarks.
On SatBird, all variants trail GeoCLIP and AlphaEarth (e.g., MS-L40: 0.558 vs.\ 0.632 and 0.675), consistent with
species distributions depending on vegetation structure, canopy cover, and microclimate — signals not explicitly
represented in OSM but directly observable from satellite data. A similar pattern appears in wildfire forecasting,
where OSMGraphCLIP trails AlphaEarth and GeoCLIP modestly, likely reflecting the importance of vegetation load and
terrain characteristics for fire risk estimation.

GeoCLIP and GT-Loc perform particularly well on geographically clustered benchmarks, such as California housing
and several health outcomes, likely because their GPS-aligned visual pretraining implicitly captures local
built-environment characteristics. Nevertheless, OSMGraphCLIP surpasses both on the majority of health tasks,
indicating that explicit semantic representations of infrastructure and urban form can provide a stronger signal
than visually inferred proxies for many public-health outcomes.

Across benchmarks, multiscale variants consistently outperform their adaptive counterparts, while L40 generally
exceeds L10, demonstrating the value of both broader contextual aggregation through band attention and
higher-resolution spherical harmonics.

\subsection{Analysis of Location Embeddings}
\label{sec:embedding_analysis}

In this section we analyze the spatial structure of the learned location embeddings
qualitatively.

\begin{figure}[!ht]
  \centering
  \includegraphics[width=\textwidth]{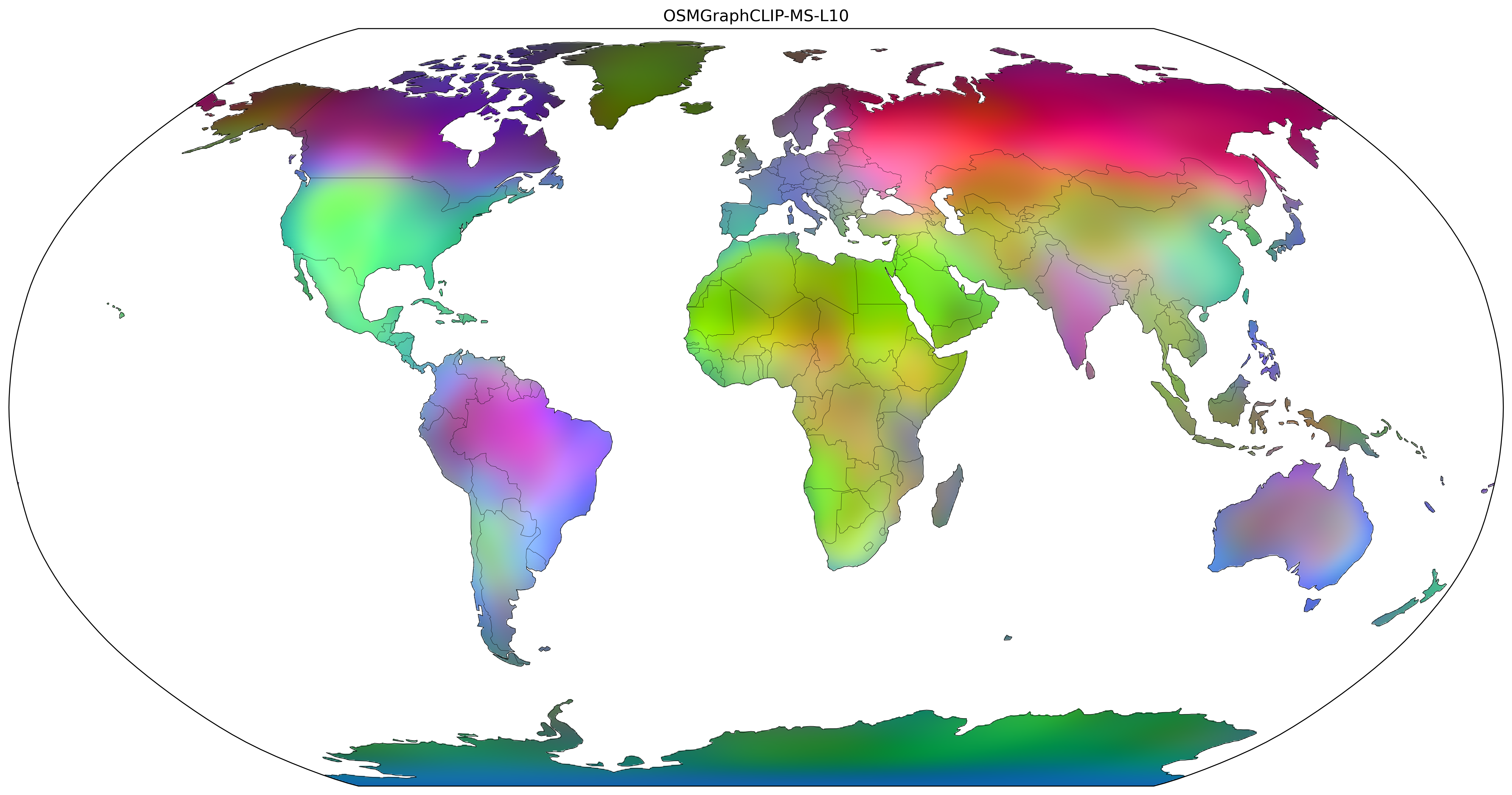}\\[2pt]
  {\small MS-L10}\\[10pt]
  \includegraphics[width=\textwidth]{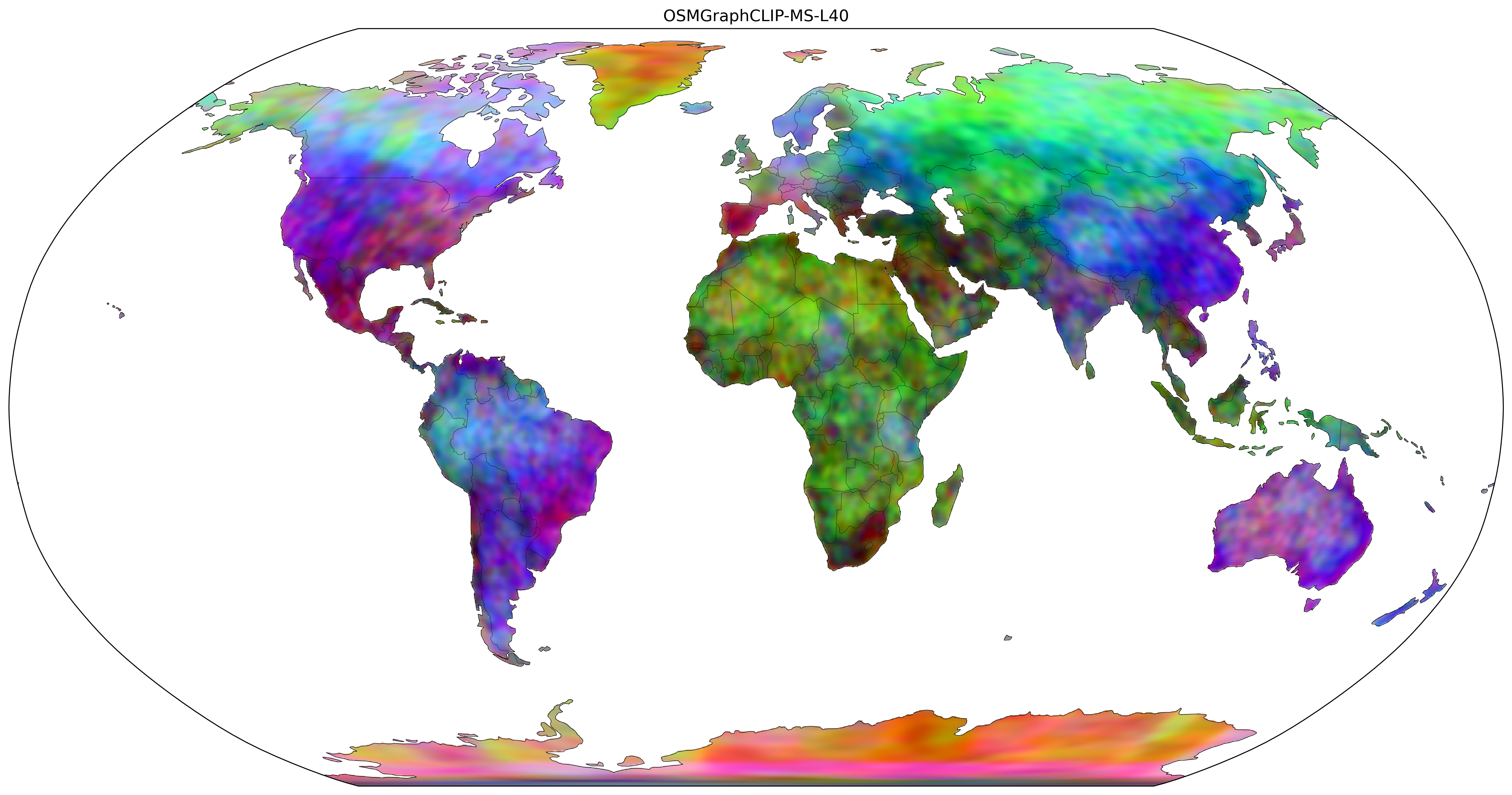}\\[2pt]
  {\small MS-L40}
  \captionsetup{skip=1pt}
  \caption{%
    RGB visualization of the first three principal components of
    OSMGraphCLIP location embeddings computed on a global grid, for MS-L10 and MS-L40
    models.
    PCA is computed independently per model; colors are therefore not
    comparable.}
  \label{fig:pca_maps}
\end{figure}

\paragraph{PCA visualization.}
Figure~\ref{fig:pca_maps} shows an RGB rendering of the first three principal
components of the OSMGraphCLIP MS-L40 and MS-L10 location encoders, evaluated
on a global grid and projected from the 256-dimensional embedding space.
The embeddings segment the globe into
spatially coherent regions: temperate forests, tropical belts, arid zones, and
dense urban corridors emerge as distinct color regions without any explicit
geographic supervision.
The L40 variant exhibit noticeably finer spatial resolution than the L10
counterpart, consistent with higher-degree spherical harmonics encoding more
localized coordinate variation.

The multiscale models shown in Figure~\ref{fig:pca_maps} produce smooth,
structured global color fields, with coherent transitions across major
geographic regions.
Boundaries such as the Sahara--Sahel transition, the boreal forest belt,
and the Indo-Gangetic Plain remain clearly delineated, indicating that the
learned embeddings capture geographically consistent large-scale structure.
See Appendix~\ref{app:embedding_analysis} for
additional visualizations and analyzes.

\begin{figure}[!ht]
  \centering
  \includegraphics[width=0.85\textwidth]{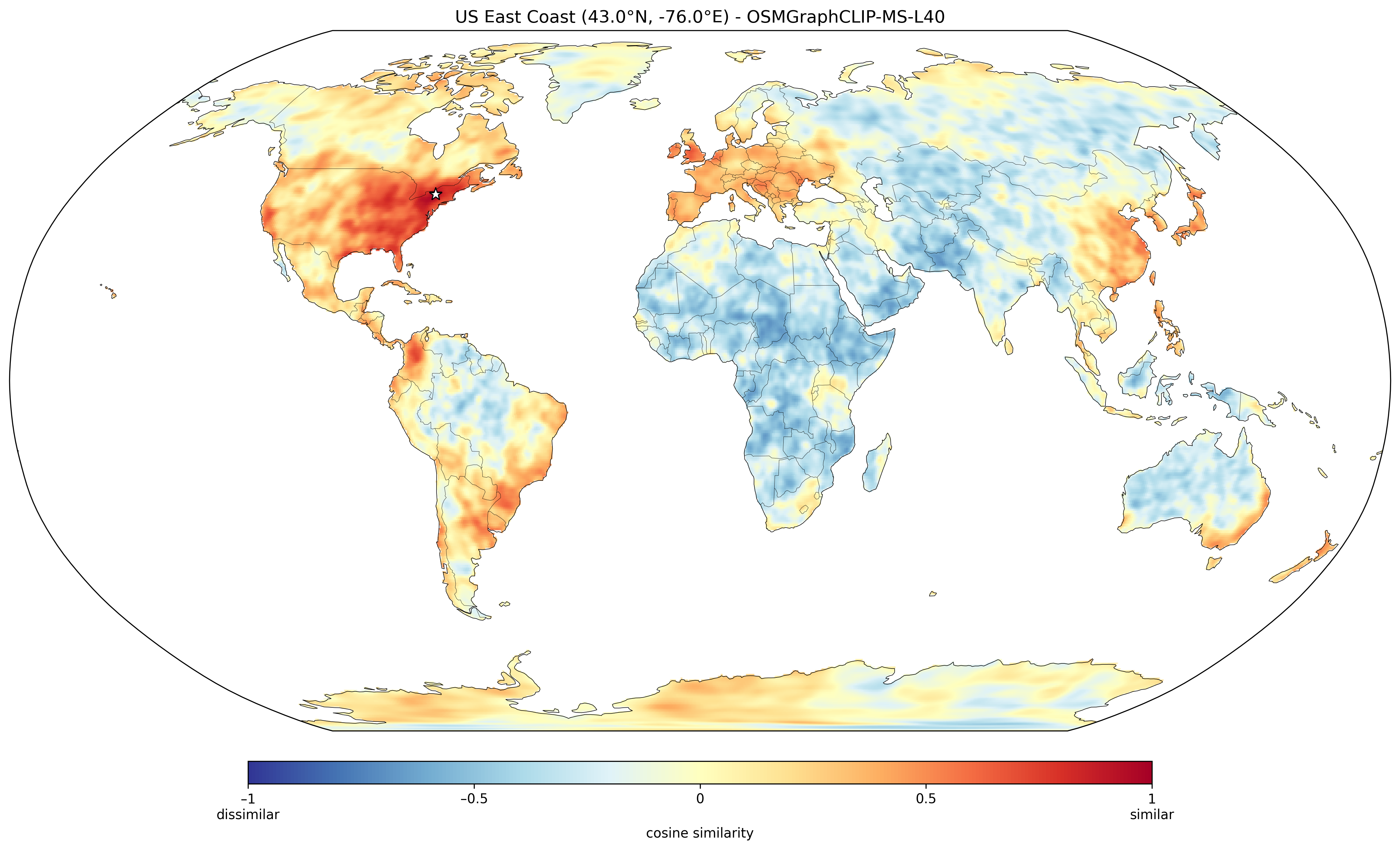}\\[2pt]
  {\small US East Coast --- MS-L40}\\[10pt]
  \includegraphics[width=0.85\textwidth]{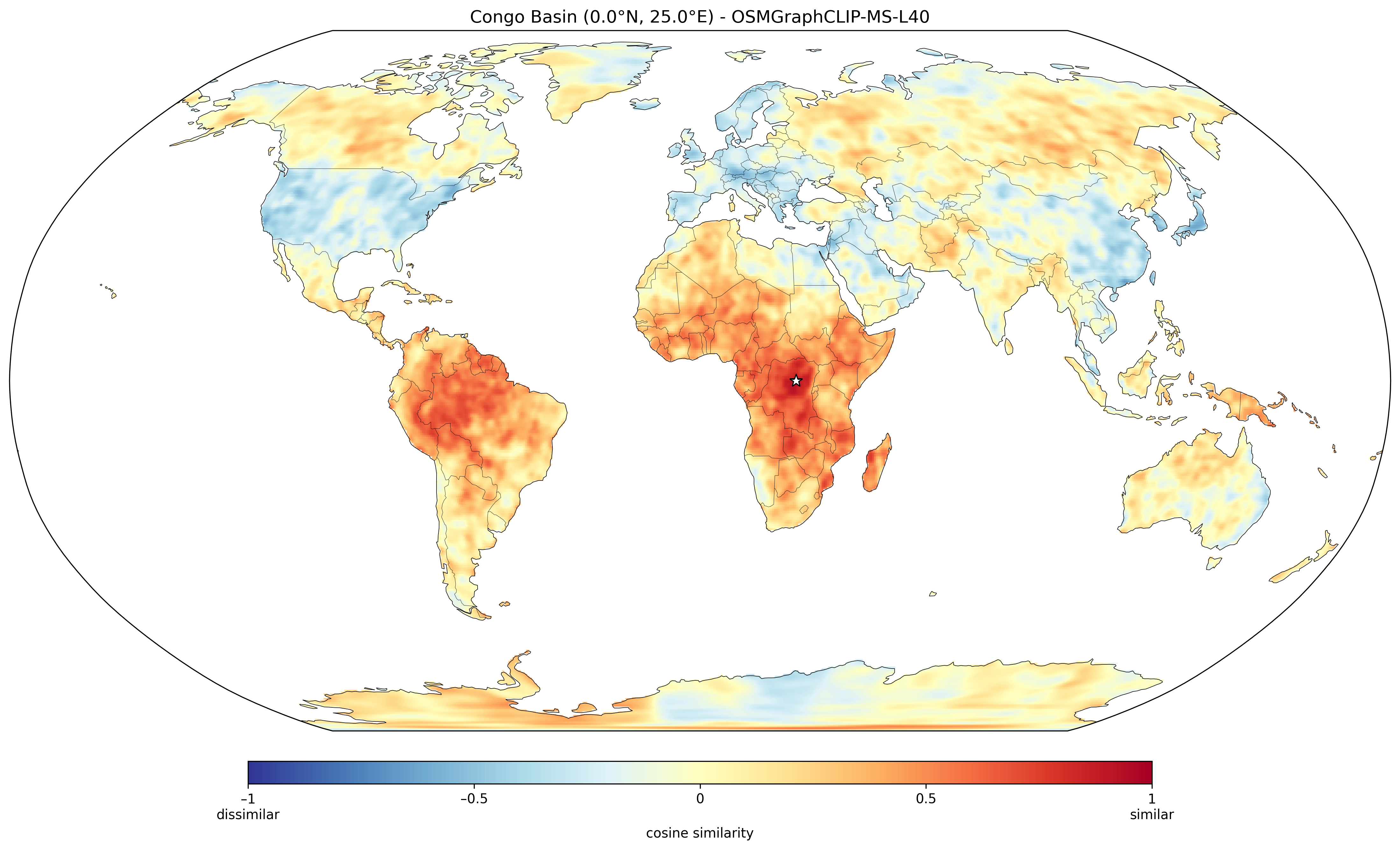}\\[2pt]
  {\small Congo Basin --- MS-L40}
  \captionsetup{skip=1pt}
  \caption{%
    Cosine similarity between two reference locations (marked $\star$) and
    all other locations on a global grid, for the MS-L40 model.
    \emph{Top}: US East Coast reference, associated with densely urbanized
    and commercially developed coastal areas in Western Europe and Northeast
    Asia.
    \emph{Bottom}: Congo Basin reference, associated with equatorial forested and
    wetland regions (Amazon, West Africa, insular Southeast Asia).
  }
  \label{fig:sim_maps}
\end{figure}

\paragraph{Cosine similarity analysis.}
Figure~\ref{fig:sim_maps} shows the cosine similarity (produced using the MS-L40 model)
between the location embedding of two reference points --- a site
on the US East Coast and a site in the Congo Basin --- and all other locations
on a global grid. We select
similar reference points as in Klemmer et al.~\cite{klemmer2025satclip} to demonstrate
the effectiveness of the learned representations across geographically and environmentally
distinct regions.

The Congo Basin reference exhibits high similarity with other
equatorial locations characterized by dense natural vegetation and low
human development: the Amazon basin, Central and West Africa, and parts of
insular Southeast Asia score highly.
This pattern reflects OSM's tagging of tropical cover
(\texttt{natural:forest}, \texttt{natural:wetland}) and the relative absence
of built-environment features across these regions.

The US East Coast reference is most similar to other
high-density urban and peri-urban areas: Western Europe, the Northeast
Asian coast (Korea, Japan, eastern China), and coastal Australia.
These regions share dense road networks, high amenity coverage, and
commercial land-use patterns --- the OSM semantic vocabulary that directly
encodes urban economic character.

\subsection{Discussion}
\label{sec:discussion}

The results reveal a modality-dependent pattern rather than a uniform ranking
across all benchmarks. OSMGraphCLIP is strongest when the prediction target is closely related to human-defined
geographic semantics and the organization of the built environment. This is most evident in the public-health
and socioeconomic benchmarks, where the best OSMGraphCLIP variants are highly competitive with, and often outperform,
image-based or Earth-observation-based baselines. In particular, MS-L40 achieves the best result on seven of the
twelve CDC PLACES outcomes, while A-L40 obtains the strongest performance on median income. These results suggest
that OSM-derived supervision captures information that is directly relevant to social, health, and urban-function
prediction tasks: road hierarchy, amenity structure, land-use function, settlement density, and the topological
organization of geographic entities.

This strength follows naturally from the modality used during pre-training. Satellite and ground-level imagery only indirectly encode the functional role of geographic entities and the semantic organization of the built environment. OSMGraphCLIP, by contrast, learns directly from what a place is used for and how
its geographic entities are relationally organized. A satellite image may reveal that a location contains buildings,
roads, or green areas, but OSM can explicitly encode whether these correspond to hospitals, restaurants, schools,
motorways, residential districts, industrial zones, parks, or other functional categories. Moreover, by representing
OSM features as a heterogeneous graph rather than rasterizing or aggregating them into feature counts, OSMGraphCLIP
preserves spatial relations such as containment, adjacency, crossing, and connectivity. This semantic-topological
signal appears particularly useful for tasks where human activity and urban structure are more predictive than
spectral signal alone.

The complementary nature of these representations is also visible in the tasks where OSMGraphCLIP is less competitive.
On habitat- and vegetation-sensitive tasks, such as SatBird and wildfire forecasting, imagery- and Earth-observation-based
models retain clear advantages. These tasks depend strongly on vegetation structure, canopy cover, terrain, microclimate,
fuel load, and seasonal conditions, i.e., signals that are directly observable in satellite data but only indirectly or
sparsely represented in OSM. Similarly, AlphaEarth performs particularly strongly on elevation and air temperature,
consistent with the value of multimodal Earth-observation data for physical-environmental prediction. These observations
indicate that OSM graphs and visual Earth-observation modalities encode different, complementary views of geographic space.

Importantly, even on those tasks where imagery- and Earth-observation-based models hold an advantage, OSMGraphCLIP
variants remain highly competitive in absolute terms --- for instance, MS-L40 matches SatCLIP-L40 on wildfire
average precision (0.794 vs.\ 0.790) and closely approaches the best coordinate-based methods on biome and ecoregion
classification --- demonstrating the breadth of the signal encoded in structured map data.

The spatial structure of the learned embeddings (Figures~\ref{fig:pca_maps}--\ref{fig:sim_maps}) supports
this interpretation: the representations organize geographic space by semantic function and environmental
character rather than geographic proximity alone, recovering biome-like gradients and separating urban
from non-urban regions without any satellite supervision.

A boundary of the current approach is the geographic unevenness of OSM coverage itself.
Regions with long-established mapping communities and high volunteer activity --- including much of North America,
Western Europe, and East Asia --- benefit from dense, fine-grained annotation, whereas areas where community
engagement, internet access, or mapping focus have historically been lower --- including parts of sub-Saharan
Africa, Central Asia, and the Amazon basin --- remain sparsely mapped.
Both model families incorporate partial mitigations: the adaptive variants (OSMGraphCLIP-A) widen the query
bounding box progressively in data-sparse locations, while the multiscale variants (OSMGraphCLIP-MS) draw on
coarser band context at broader radii where fine-grained features are absent.
Neither mechanism can recover semantic richness that volunteers have not yet contributed, however.
Representations for well-annotated regions are therefore likely richer than those for sparsely mapped environments,
something that practitioners should keep in mind for data-scarce geographies.

\section{Conclusions}
\label{sec:conclusions}

We have presented OSMGraphCLIP, a geospatial representation model that
learns globally transferable location embeddings by contrastively aligning
heterogeneous OSM graphs --- and coarser multi-scale band context --- with
a spherical-harmonics location encoder.
Evaluated across 24 downstream tasks spanning climate, ecology, land cover,
biodiversity, socioeconomics, public health, and wildfire forecasting,
OSMGraphCLIP demonstrates that structured, collaboratively curated map data
constitutes a powerful supervisory modality for global location
representation learning.
The model achieves state-of-the-art or near-state-of-the-art performance on
the majority of benchmarks, with particular strength on tasks where the
semantic organization of the built environment is the dominant predictive
signal, and it remains highly competitive even on tasks where imagery-based
models hold an advantage, underscoring the breadth of the geographic signal
encoded in OSM graph topology.

Overall, these results position OSMGraphCLIP as a complementary
semantic-topological location representation model. Structured,
collaboratively curated map data constitutes a viable, and in some domains
superior, supervisory modality for global location representation learning,
especially where explicit place function and spatial topology matter.
The most impactful next steps are tri-modal contrastive training that aligns
OSM graphs, satellite imagery, and geographic coordinates within a shared
embedding space, and exploiting the temporal versioning of OSM to learn
representations of urban change.

\section*{Acknowledgements}
This work has received funding from the European Union's Horizon Europe WIDERA
Coordination and Support Actions under Grant Agreement no.101159723 (MeDiTwin).

During the preparation of this work, the authors used generative AI tools for
language editing and text refinement. The authors take full responsibility for
the content of this work.

\clearpage
\appendix
\section{Appendix}

\subsection{Evaluation Protocol Details}
\label{app:Evaluation}

\subsubsection{Dataset Overview}

Unless otherwise specified, we use official benchmark splits and preprocessing protocols. For California Housing we use the standard \texttt{scikit-learn} implementation.
For Median Income, we construct a county-level dataset from USDA Economic Research Service 2022 median household
income estimates and assign representative coordinates using U.S.\ Census county boundaries. For CDC PLACES, each ZCTA is mapped to a centroid coordinate which is given in the dataset. For SatBird, reBEN, and Mesogeos we use the official
benchmark splits. In Mesogeos, samples are represented by the latitude--longitude coordinate pair of the corresponding grid-cell centroid. Table~\ref{tab:location-encoder-benchmark-tasks}
summarises benchmark datasets, split configurations,
evaluation metrics, and geographic coverage.

\begin{table*}[!t]
\centering
\small
\caption{Summary of benchmark evaluation tasks. For iNaturalist, 10\% of
the official training split is used for validation, and the official
validation split is used as the test set. CDC PLACES tasks share the same
configuration and are grouped for brevity.}
\label{tab:location-encoder-benchmark-tasks}
\vspace{0.5em}
\renewcommand{\arraystretch}{1.20}
\begin{tabular}{
p{3.6cm}
>{\centering\arraybackslash}p{1.5cm}
>{\centering\arraybackslash}p{1.4cm}
>{\centering\arraybackslash}p{1.2cm}
>{\centering\arraybackslash}p{1.4cm}
>{\centering\arraybackslash}p{1.6cm}
}
\toprule
\multicolumn{1}{c}{\textbf{Task name}} &
\multicolumn{1}{c}{\textbf{Source}} &
\multicolumn{1}{c}{\textbf{Samples}} &
\multicolumn{1}{c}{\textbf{Split}} &
\multicolumn{1}{c}{\textbf{Metric}} &
\multicolumn{1}{c}{\textbf{Coverage}} \\
\midrule

\multicolumn{6}{@{}l}{\textbf{Regression}} \\

Air temperature~\cite{hooker2018airtemp}
& \multirow{5}{*}{SatCLIP}
& 3{,}076 & 60-20-20 & $R^2$ & Global \\

Median income~\cite{jia2020residual}
& 
& 3{,}219 & 50-20-30 & $R^2$ & Cont.\ USA \\

California housing~\cite{pace2003semiparametric}
& 
& 20{,}640 & 50-20-30 & $R^2$ & Calif., USA \\

Elevation~\cite{rolf2021generalizable}
& 
& 100{,}000 & 30-10-60 & $R^2$ & Global \\

Population~\cite{rolf2021generalizable}
& 
& 74{,}512 & 30-10-60 & $R^2$ & Global \\

\midrule

SatBird -- USA summer~\cite{satbird_NEURIPS2023}
& SatBird
& 122{,}593 & Predefined & Top-$k$ Acc. & USA \\

\midrule

CDC PLACES (12 outcomes)~\cite{cdcplaces2023}
& CDC PLACES
& 32{,}409 & 80-10-10 & $R^2$ & Cont.\ USA \\

\midrule

\multicolumn{6}{@{}l}{\textbf{Classification}} \\

Countries~\cite{klemmer2025satclip}
& \multirow{4}{*}{SatCLIP}
& 100{,}000 & 30-10-60 & Accuracy & Global \\

iNaturalist~\cite{horn2018inaturalist}
& 
& 460{,}406 & Predefined & Accuracy & Global \\

Biome~\cite{dinerstein2017ecoregion}
& 
& 100{,}000 & 30-10-60 & Accuracy & Global \\

Ecoregions~\cite{dinerstein2017ecoregion}
& 
& 100{,}000 & 30-10-60 & Accuracy & Global \\

\midrule

reBEN~\cite{clasen2024reben}
& BigEarthNet
& 549{,}488 & Predefined & micro-F1 & Europe \\

\midrule

Wildfire forecasting~\cite{kondylatos2023mesogeos}
& Mesogeos
& 16{,}242 & Predefined & AUPRC & Mediterranean \\

\bottomrule
\end{tabular}
\end{table*}

\subsubsection{Embedding Extraction}

For each task, frozen embeddings are extracted from
each encoder and used as input to downstream predictors.
The embedding dimensionality for each encoder is:
SatCLIP~\cite{klemmer2025satclip} (256),
GeoCLIP~\cite{geoclip2023} (512),
\textsc{GT-Loc}~\cite{shatwell2025gt} (512),
AlphaEarth Foundations (AEF)~\cite{brown2025alphaearth} (64),
Copernicus-FM~\cite{wang2025unifiedcopernicusfoundationmodel} (768),
and OSMGraphCLIP (ours) (256).

\subsubsection{Embedding Extraction}

For each task, frozen embeddings are extracted from
each encoder and used as input to downstream predictors.
Table~\ref{tab:encoder-embedding-dimensions}
reports embedding dimensionality.

\begin{table}[!t]
\centering
\small
\caption{Embedding dimensionality for each evaluated encoder.}
\label{tab:encoder-embedding-dimensions}
\vspace{0.5em}
\setlength{\tabcolsep}{6pt}
\renewcommand{\arraystretch}{1.35}
\begin{tabular}{
p{5.2cm}
>{\centering\arraybackslash}p{2.5cm}
}
\toprule
\textbf{Encoder} & \textbf{Dimension} \\
\midrule
SatCLIP~\cite{klemmer2025satclip}                 & 256 \\
GeoCLIP~\cite{geoclip2023}                       & 512 \\
\textsc{GT-Loc}~\cite{shatwell2025gt}            & 512 \\
AlphaEarth Foundations (AEF)~\cite{brown2025alphaearth} & 64 \\
Copernicus-FM~\cite{wang2025unifiedcopernicusfoundationmodel} & 768 \\
OSMGraphCLIP (ours)                                 & 256 \\
\bottomrule
\end{tabular}
\end{table}

\paragraph{Copernicus-FM}
We use the publicly released Copernicus-Embed-025deg grid, which provides
a global embedding map at $0.25^\circ$ resolution with shape
$721 \times 1440 \times 768$.
Each grid cell contains a 768-dimensional embedding obtained by averaging
Copernicus-FM representations across modalities.
For each downstream coordinate pair we map to the corresponding
$0.25^\circ$ grid cell and retrieve the embedding directly.

\paragraph{AlphaEarth Foundations (AEF)}
We use the 2023 annual embedding field from the released Google Earth Engine
dataset.
AEF provides analysis-ready embedding fields over Earth's terrestrial
surface with 64 embedding bands per pixel at 10\,m resolution.
We sample the 2023 image at each downstream coordinate using Google Earth
Engine.
Since some coordinates may fall on masked pixels or outside valid AEF
coverage, we first attempt direct point sampling and then apply a
nearest-valid-pixel fallback: for missing points we search within
increasing buffer radii up to 10\,km and use the closest valid embedding
if one is found.
Coordinates for which no valid embedding is available within this radius
are treated as missing.
Because AEF covers terrestrial surfaces only, it cannot produce embeddings
for ocean locations; we therefore exclude AEF from the Country, Biome, and
Ecoregion tasks and report N/A for these entries.
Missing AEF entries are excluded from aggregate win-count statistics and
are not penalized as failures.

\subsubsection{Downstream Predictors}

For each task and encoder we train a two-layer MLP with hidden dimension 128 and dropout rate 0.5.
All models are trained using the AdamW optimizer with learning rate $10^{-3}$
and weight decay $10^{-5}$, using a batch size of 1{,}024. Unless otherwise specified, training is performed for 100 epochs and repeated over ten random seeds (0, 1, 7, 11, 42, 100, 1234, 2021, 8657, 41674), with performance reported as the mean over runs. Due to computational cost, SatBird is trained for 20 epochs using three seeds (0, 16, 1234).
Table~\ref{tab:downstream-training-hyperparameters} summarises training hyperparameters.

\begin{table}[!t]
\centering
\small
\caption{Downstream training hyperparameters used for all tasks unless
otherwise specified.}
\label{tab:downstream-training-hyperparameters}
\vspace{0.5em}
\setlength{\tabcolsep}{6pt}
\renewcommand{\arraystretch}{1.35}
\begin{tabular}{
p{5.0cm}
>{\centering\arraybackslash}p{2.8cm}
}
\toprule
\textbf{Hyperparameter} & \textbf{Value} \\
\midrule
Epochs                  & 100            \\
Batch size              & 1{,}024        \\
Optimizer               & AdamW          \\
Learning rate           & $10^{-3}$      \\
Weight decay            & $10^{-5}$      \\
MLP hidden dimension    & 128            \\
MLP number of layers    & 2              \\
Dropout                 & 0.5            \\
Evaluation models       & MLP \\
Seeds & \begin{tabular}{@{}c@{}}0, 1, 7, 11, 42, 100,\\ 1234, 2021, 8657, 41674\end{tabular} \\
\bottomrule
\end{tabular}
\end{table}

\begin{figure}[!htb]
  \centering
  \includegraphics[width=\columnwidth]{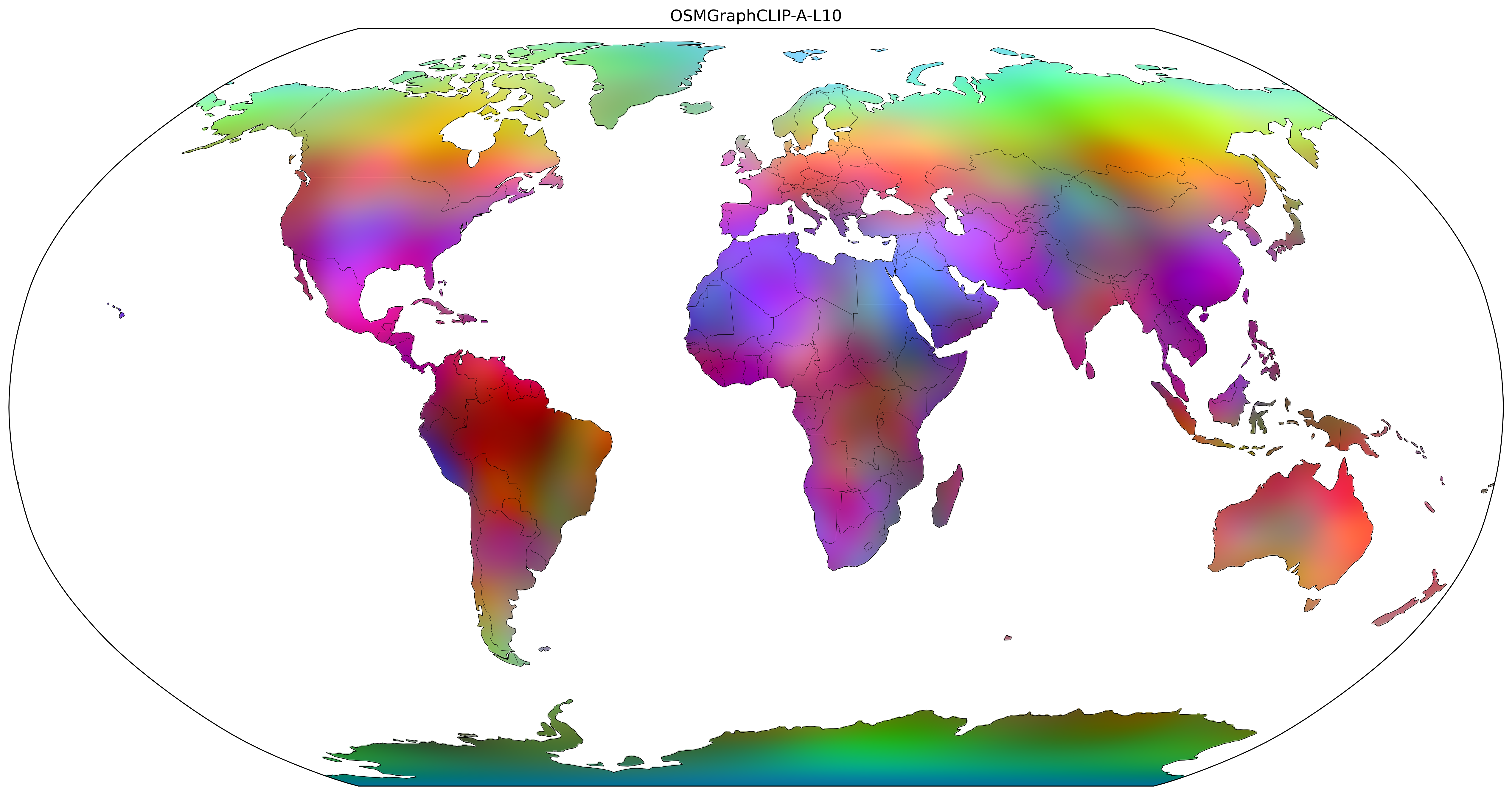}\\[2pt]
  {\small A-L10}\\[6pt]
  \includegraphics[width=\columnwidth]{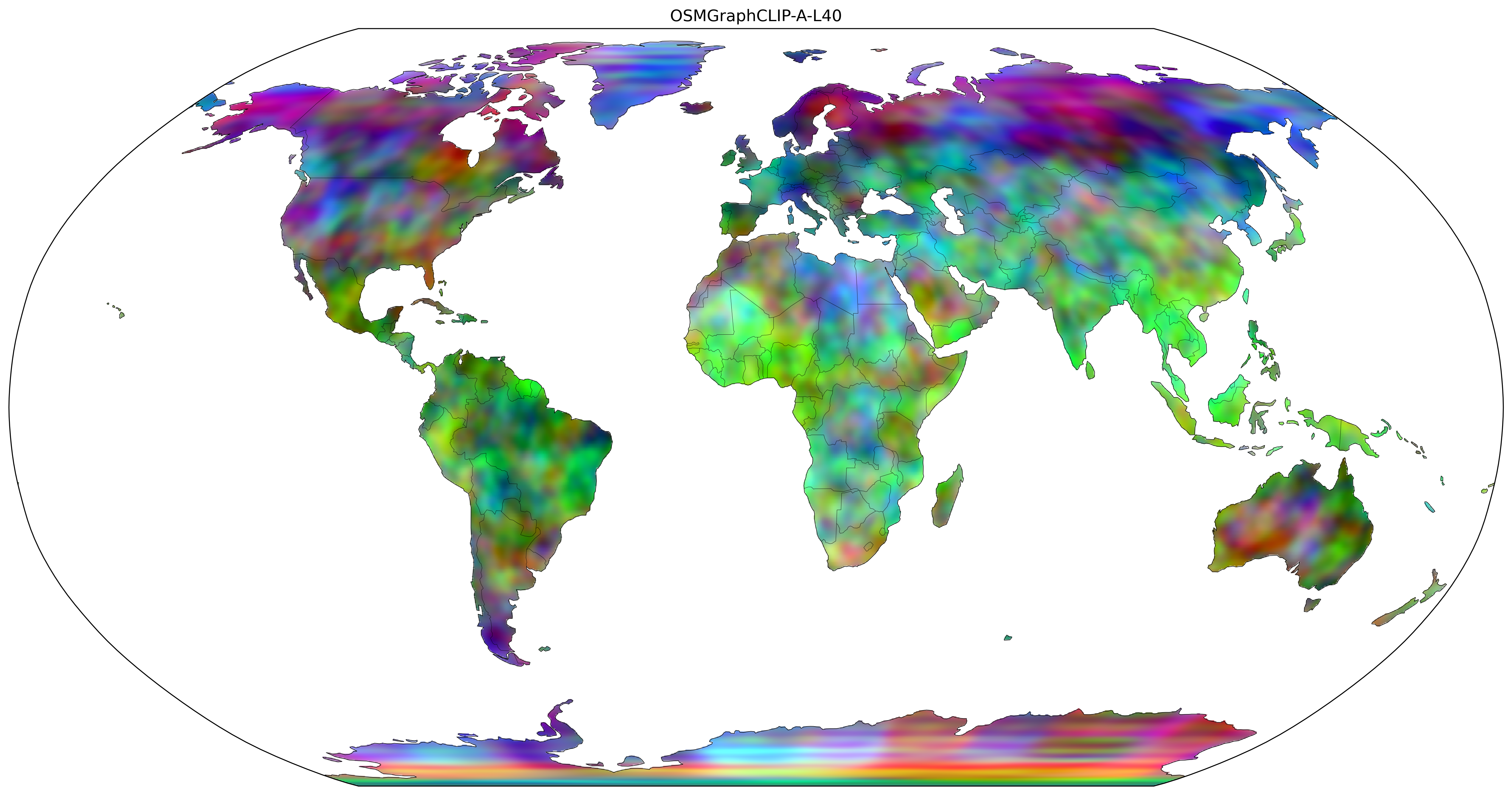}\\[2pt]
  {\small A-L40}
  \caption{%
    RGB visualization of the first three principal components of
    OSMGraphCLIP location embeddings computed on a global grid, for A-L10 and A-L40
    model variants.
    PCA is computed independently per model; colors are therefore not
    comparable across globes.
  }
  \label{fig:pca_maps_adaptive}
\end{figure}

\begin{figure*}[!t]
  \centering
  \includegraphics[width=0.85\textwidth]{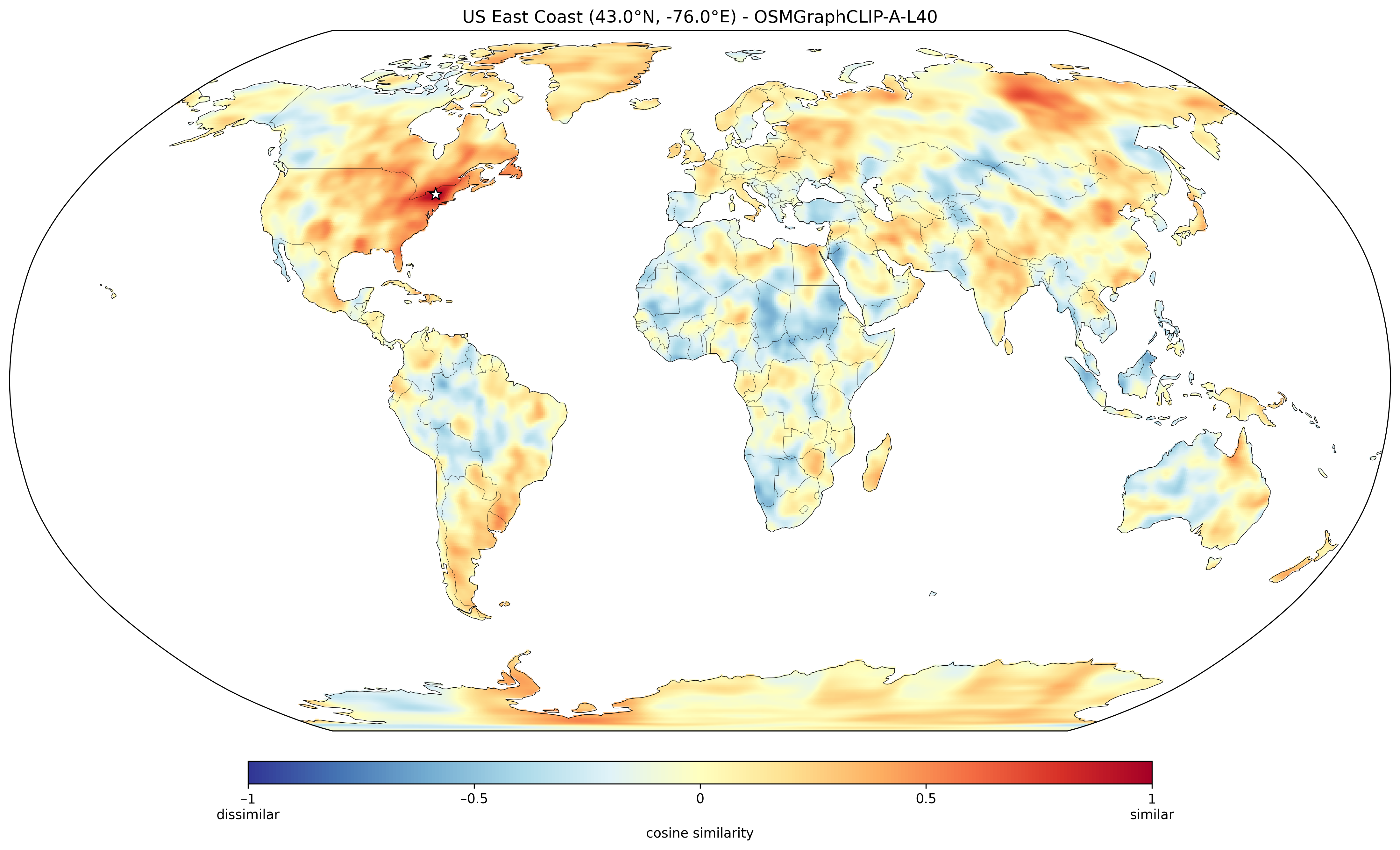}\\[2pt]
  {\small US East Coast --- A-L40}\\[10pt]
  \includegraphics[width=0.85\textwidth]{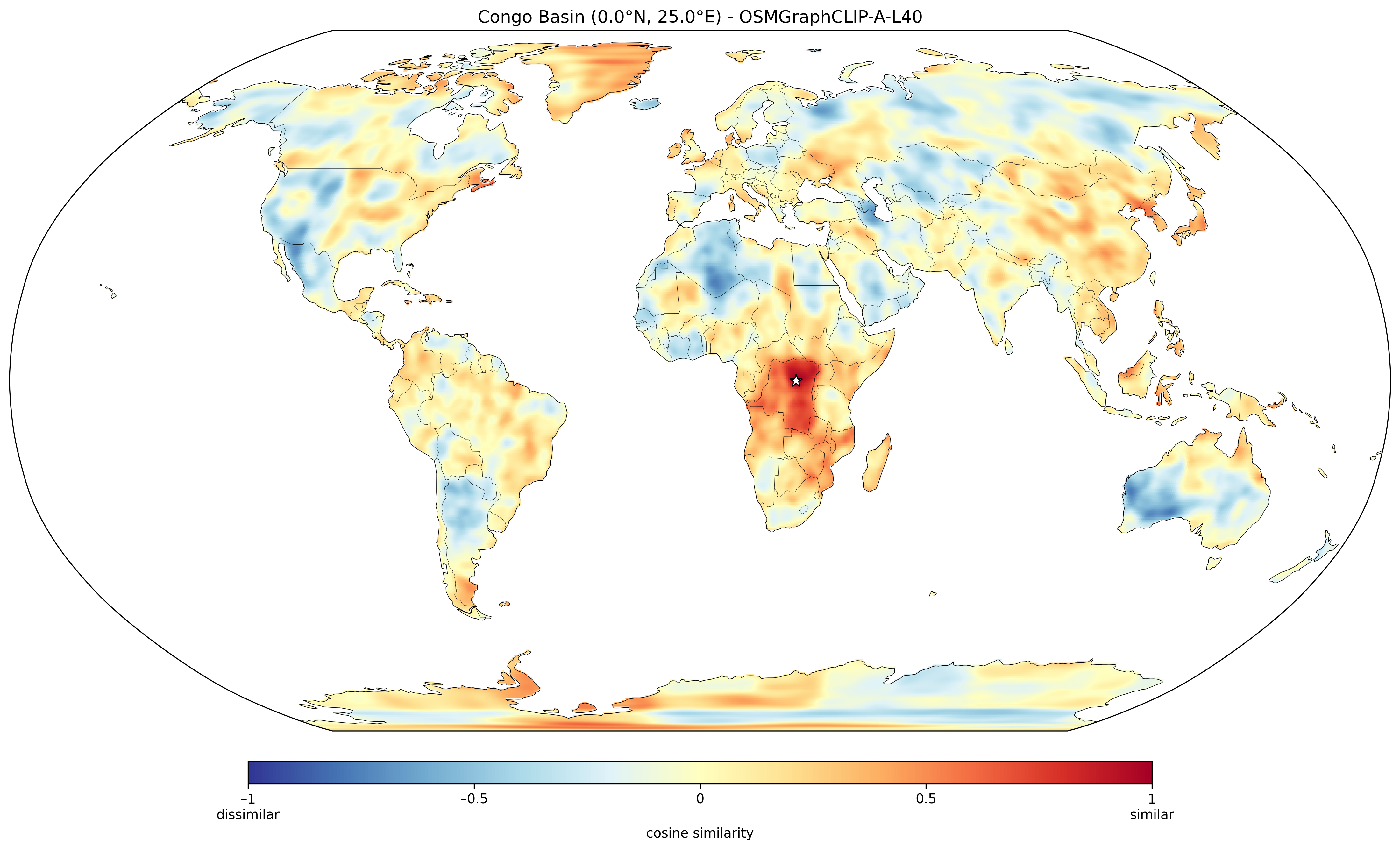}\\[2pt]
  {\small Congo Basin --- A-L40}
  \caption{%
    Cosine similarity between two reference locations (marked $\star$) and
    all other locations on a global grid, for the A-L40 adaptive model.
  }
  \label{fig:sim_maps_adaptive}
\end{figure*}

\subsubsection{Task-Specific Inputs}

\paragraph{iNaturalist.}
Following the SatCLIP evaluation protocol~\cite{klemmer2025satclip} and
the geo-prior evaluation of Mac Aodha et al.~\cite{mac2019presence}, the
location embedding is concatenated with pretrained InceptionV3 image
features before training the species classifier.
This is the only task for which image-derived features are used.

\paragraph{Wildfire danger forecasting.}
Following the Mesogeos protocol~\cite{kondylatos2023mesogeos}, a cyclic
day-of-year encoding is concatenated with the location embedding to capture seasonality:
\[
    \sin\!\left(2\pi \tfrac{d}{365}\right), \qquad
    \cos\!\left(2\pi \tfrac{d}{365}\right),
\]
where $d$ denotes the day of year extracted from the Mesogeos sample
metadata.
These two temporal features are concatenated with the location embedding
before training the downstream predictor.
This is the only task for which temporal features are used.

\subsection{PCA Visualization of Location Embeddings}
\label{app:embedding_analysis}

Figure~\ref{fig:pca_maps_adaptive} shows RGB visualizations of the first three principal
components of OSMGraphCLIP location embeddings computed on a global grid, for A-L10 and 
A-L40 model variants.

\subsection{Cosine Similarity Maps}
\label{app:similarity_maps}

Figure~\ref{fig:sim_maps_adaptive} shows cosine similarity maps for the A-L40 adaptive model.


\clearpage

\begin{thebibliography}{40}
\providecommand{\natexlab}[1]{#1}
\providecommand{\url}[1]{\texttt{#1}}
\expandafter\ifx\csname urlstyle\endcsname\relax
  \providecommand{\doi}[1]{doi: #1}\else
  \providecommand{\doi}{doi: \begingroup \urlstyle{rm}\Url}\fi

\bibitem[Agarwal et~al.(2024)Agarwal, Sun, Kamath, Muslim, Sarker, Paul, Yee,
  Sieniek, Jablonski, Mayer, et~al.]{agarwal2024geospatial}
Mohit Agarwal, Mimi Sun, Chaitanya Kamath, Arbaaz Muslim, Prithul Sarker,
  Joydeep Paul, Hector Yee, Marcin Sieniek, Kim Jablonski, Yael Mayer, et~al.
\newblock General geospatial inference with a population dynamics foundation
  model.
\newblock \emph{arXiv preprint arXiv:2411.07207}, 2024.

\bibitem[Authors(2025{\natexlab{a}})]{gair2025}
GAIR Authors.
\newblock {GAIR}: Aligning satellite, street view, and location embeddings via
  contrastive learning.
\newblock \emph{arXiv preprint arXiv:2503.16683}, 2025{\natexlab{a}}.

\bibitem[Authors(2025{\natexlab{b}})]{h3mosaic2025}
H3-MOSAIC Authors.
\newblock {H3-MOSAIC}: Combining {OSM} semantics and satellite imagery on
  spatial grids.
\newblock \emph{International Journal of Health Geographics},
  2025{\natexlab{b}}.

\bibitem[Bai et~al.(2025)Bai, Zhang, Zhang, Zhang, Wang, Qin, and
  Du]{bai2025geolink}
Lubian Bai, Xiuyuan Zhang, Siqi Zhang, Zepeng Zhang, Haoyu Wang, Wei Qin, and
  Shihong Du.
\newblock Geolink: Empowering remote sensing foundation model with
  openstreetmap data.
\newblock \emph{arXiv preprint arXiv:2509.26016}, 2025.

\bibitem[Brodsky(2018)]{h3uber2018}
Isaac Brodsky.
\newblock {H3}: {Uber}'s hexagonal hierarchical spatial index.
\newblock Uber Engineering Blog, 2018.
\newblock URL \url{https://eng.uber.com/h3/}.
\newblock Accessed 2026.

\bibitem[Brown et~al.(2025)Brown, Kazmierski, Pasquarella, Rucklidge,
  Samsikova, Zhang, Shelhamer, Lahera, Wiles, Ilyushchenko,
  et~al.]{brown2025alphaearth}
Christopher~F Brown, Michal~R Kazmierski, Valerie~J Pasquarella, William~J
  Rucklidge, Masha Samsikova, Chenhui Zhang, Evan Shelhamer, Estefania Lahera,
  Olivia Wiles, Simon Ilyushchenko, et~al.
\newblock Alphaearth foundations: An embedding field model for accurate and
  efficient global mapping from sparse label data.
\newblock \emph{arXiv preprint arXiv:2507.22291}, 2025.

\bibitem[{Centers for Disease Control and Prevention}(2023)]{cdcplaces2023}
{Centers for Disease Control and Prevention}.
\newblock {PLACES}: Local data for better health, {ZCTA} data ({GIS}-friendly
  format), 2023 release.
\newblock Data.CDC.gov, 2023.
\newblock URL
  \url{https://data.cdc.gov/500-Cities-Places/PLACES-ZCTA-Data-GIS-Friendly-Format-2023-release/c7b2-4ecy/about_data}.
\newblock Accessed 2026.

\bibitem[Clasen et~al.(2024)Clasen, Hackel, Burgert, Sumbul, Demir, and
  Markl]{clasen2024reben}
Kai~Norman Clasen, Leonard Hackel, Tom Burgert, Gencer Sumbul, Beg{\"u}m Demir,
  and Volker Markl.
\newblock r{eBEN}: Refined {BigEarthNet} dataset for remote sensing image
  analysis.
\newblock \emph{arXiv preprint arXiv:2407.03653}, 2024.

\bibitem[Clementini et~al.(1993)Clementini, Di~Felice, and
  Van~Oosterom]{clementini1993small}
Eliseo Clementini, Paolino Di~Felice, and Peter Van~Oosterom.
\newblock A small set of formal topological relationships suitable for end-user
  interaction.
\newblock In \emph{International symposium on spatial databases}, pages
  277--295. Springer, 1993.

\bibitem[de~Almeida et~al.(1998)de~Almeida, Raper, Camara, and
  Cova]{de1998spatial}
Jo{\~{a}}o~Paulo de~Almeida, Jonathan Raper, Gilberto Camara, and Thomas Cova.
\newblock A formal approach to imprecise and incomplete geographical objects.
\newblock \emph{Computers, Environment and Urban Systems}, 22\penalty0
  (5):\penalty0 395--408, 1998.

\bibitem[Dinerstein et~al.(2017)Dinerstein, Olson, Joshi, Vynne, Burgess,
  Wikramanayake, Hahn, Palminteri, Hedao, Noss,
  et~al.]{dinerstein2017ecoregion}
Eric Dinerstein, David Olson, Anup Joshi, Carly Vynne, Neil~D Burgess, Eric
  Wikramanayake, Nathan Hahn, Suzanne Palminteri, Prashant Hedao, Reed Noss,
  et~al.
\newblock An ecoregion-based approach to protecting half the terrestrial realm.
\newblock \emph{BioScience}, 67\penalty0 (6):\penalty0 534--545, 2017.

\bibitem[Donghi and Morvan(2023)]{donghi2023geovex}
Daniele Donghi and Anne Morvan.
\newblock Geovex: Geospatial vectors with hexagonal convolutional autoencoders.
\newblock In \emph{Proceedings of the 6th ACM SIGSPATIAL International Workshop
  on AI for Geographic Knowledge Discovery}, pages 3--13, 2023.

\bibitem[Hooker et~al.(2018)Hooker, Duveiller, and Cescatti]{hooker2018airtemp}
Jake Hooker, Gregory Duveiller, and Alessandro Cescatti.
\newblock A global dataset of air temperature derived from satellite remote
  sensing and weather stations.
\newblock \emph{Scientific Data}, 5\penalty0 (1):\penalty0 180246, 2018.

\bibitem[Jia and Benson(2020)]{jia2020residual}
Junteng Jia and Austin~R Benson.
\newblock Residual correlation in graph neural network regression.
\newblock In \emph{Proceedings of the 26th ACM SIGKDD International Conference
  on Knowledge Discovery \& Data Mining}, pages 588--598, 2020.

\bibitem[Klemmer et~al.(2025)Klemmer, Rolf, Robinson, Mackey, and
  Ru{\ss}wurm]{klemmer2025satclip}
Konstantin Klemmer, Esther Rolf, Caleb Robinson, Lester Mackey, and Marc
  Ru{\ss}wurm.
\newblock Satclip: Global, general-purpose location embeddings with satellite
  imagery.
\newblock In \emph{Proceedings of the AAAI Conference on Artificial
  Intelligence}, volume~39, pages 4347--4355, 2025.

\bibitem[Kondylatos et~al.(2023)Kondylatos, Prapas, Camps-Valls, and
  Papoutsis]{kondylatos2023mesogeos}
Spyros Kondylatos, Ioannis Prapas, Gustau Camps-Valls, and Ioannis Papoutsis.
\newblock Mesogeos: A multi-purpose dataset for data-driven wildfire modeling
  in the mediterranean.
\newblock In \emph{Thirty-seventh Conference on Neural Information Processing
  Systems Datasets and Benchmarks Track}, 2023.
\newblock URL \url{https://openreview.net/forum?id=VH1vxapUTs}.

\bibitem[Le{\'s}niara and Szyma{\'n}ski(2022)]{lesniara2022highway2vec}
Kacper Le{\'s}niara and Piotr Szyma{\'n}ski.
\newblock Highway2vec: Representing {OpenStreetMap} microregions with respect
  to their road network characteristics.
\newblock In \emph{Proceedings of the 5th ACM SIGSPATIAL International Workshop
  on AI for Geographic Knowledge Discovery}, pages 18--29, 2022.

\bibitem[Liu et~al.(2025)Liu, Wang, Cheng, and Law]{liu2025enriching}
Junyuan Liu, Xinglei Wang, Tao Cheng, and Stephen Law.
\newblock Enriching location representation with detailed semantic information.
\newblock In \emph{12th International Conference on Geographic Information
  Science (GIScience 2025)}, volume 352 of \emph{Leibniz International
  Proceedings in Informatics (LIPIcs)}, pages 3:1--3:7, 2025.
\newblock \doi{10.4230/LIPIcs.GIScience.2025.3}.

\bibitem[Loshchilov and Hutter(2017)]{loshchilov2017adamw}
Ilya Loshchilov and Frank Hutter.
\newblock Decoupled weight decay regularization.
\newblock \emph{arXiv preprint arXiv:1711.05101}, 2017.

\bibitem[Mac~Aodha et~al.(2019)Mac~Aodha, Cole, and Perona]{mac2019presence}
Oisin Mac~Aodha, Elijah Cole, and Pietro Perona.
\newblock Presence-only geographical priors for fine-grained image
  classification.
\newblock In \emph{Proceedings of the IEEE/CVF International Conference on
  Computer Vision}, pages 9596--9606, 2019.

\bibitem[Mai et~al.(2020)Mai, Janowicz, Yan, Zhu, Cai, and
  Lao]{mai2020space2vec}
Gengchen Mai, Krzysztof Janowicz, Bo~Yan, Rui Zhu, Ling Cai, and Ni~Lao.
\newblock Multi-scale representation learning for spatial feature distributions
  using grid cells.
\newblock In \emph{International Conference on Learning Representations}, 2020.

\bibitem[Mai et~al.(2023)Mai, Xuan, Lao, He, Cundy, Zhao, Gao, and
  Ermon]{mai2023sphere2vec}
Gengchen Mai, Yao Xuan, Ni~Lao, Jinmeng He, Chris Cundy, Weiming Zhao, Song
  Gao, and Stefano Ermon.
\newblock Sphere2vec: A general-purpose location representation learning over a
  spherical surface for large-scale geospatial predictions.
\newblock \emph{ISPRS Journal of Photogrammetry and Remote Sensing},
  202:\penalty0 439--462, 2023.

\bibitem[{OpenStreetMap Contributors}(2004)]{openstreetmap}
{OpenStreetMap Contributors}.
\newblock {OpenStreetMap}: The free wiki world map.
\newblock \emph{https://www.openstreetmap.org}, 2004.

\bibitem[Pace and Barry(2003)]{pace2003semiparametric}
R~Kelley Pace and Ronald~P Barry.
\newblock Semiparametric maximum likelihood estimates of spatial dependence.
\newblock \emph{Geographical Analysis}, 35\penalty0 (1):\penalty0 76--90, 2003.

\bibitem[Radford et~al.(2021)Radford, Kim, Hallacy, Ramesh, Goh, Agarwal,
  Sastry, Askell, Mishkin, Clark, et~al.]{radford2021clip}
Alec Radford, Jong~Woon Kim, Chris Hallacy, Aditya Ramesh, Gabriel Goh,
  Sandhini Agarwal, Girish Sastry, Amanda Askell, Pamela Mishkin, Jack Clark,
  et~al.
\newblock Learning transferable visual models from natural language
  supervision.
\newblock In \emph{Proceedings of the 38th International Conference on Machine
  Learning}, pages 8748--8763. PMLR, 2021.

\bibitem[Reimers and Gurevych(2019)]{reimers2019sbert}
Nils Reimers and Iryna Gurevych.
\newblock Sentence-{BERT}: Sentence embeddings using {Siamese BERT-Networks}.
\newblock \emph{arXiv preprint arXiv:1908.10084}, 2019.

\bibitem[Rolf et~al.(2021)Rolf, Proctor, Carleton, Bolliger, Shankar, Ishihara,
  Recht, and Hsiang]{rolf2021generalizable}
Esther Rolf, Jonathan Proctor, Tamma Carleton, Ian Bolliger, Vaishaal Shankar,
  Miyabi Ishihara, Benjamin Recht, and Solomon Hsiang.
\newblock A generalizable and accessible approach to machine learning with
  global satellite imagery.
\newblock \emph{Nature Communications}, 12\penalty0 (1):\penalty0 4392, 2021.

\bibitem[Ru{\ss}wurm et~al.(2024)Ru{\ss}wurm, Klemmer, Rolf, Zbinden, and
  Tuia]{russwurm2024spherical}
Marc Ru{\ss}wurm, Konstantin Klemmer, Esther Rolf, Robin Zbinden, and Devis
  Tuia.
\newblock Geographic location encoding with spherical harmonics and sinusoidal
  representation networks.
\newblock In \emph{International Conference on Learning Representations}, 2024.

\bibitem[Shatwell et~al.(2025)Shatwell, Dave, Swetha, and Shah]{shatwell2025gt}
David~G Shatwell, Ishan~Rajendrakumar Dave, Sirnam Swetha, and Mubarak Shah.
\newblock Gt-loc: Unifying when and where in images through a joint embedding
  space.
\newblock In \emph{Proceedings of the IEEE/CVF International Conference on
  Computer Vision}, pages 1--11, 2025.

\bibitem[Sitzmann et~al.(2020)Sitzmann, Martel, Bergman, Lindell, and
  Wetzstein]{sitzmann2020siren}
Vincent Sitzmann, Julien Martel, Alexander Bergman, David Lindell, and Gordon
  Wetzstein.
\newblock Implicit neural representations with periodic activation functions.
\newblock In \emph{Advances in Neural Information Processing Systems},
  volume~33, pages 7462--7473, 2020.

\bibitem[Teng et~al.(2023)Teng, Elmustafa, Akera, Bengio, Radi, Larochelle, and
  Rolnick]{satbird_NEURIPS2023}
M\'{e}lisande Teng, Amna Elmustafa, Benjamin Akera, Yoshua Bengio, Hager Radi,
  Hugo Larochelle, and David Rolnick.
\newblock Satbird: a dataset for bird species distribution modeling using
  remote sensing and citizen science data.
\newblock In A.~Oh, T.~Neumann, A.~Globerson, K.~Saenko, M.~Hardt, and
  S.~Levine, editors, \emph{Advances in Neural Information Processing Systems},
  volume~36, pages 75925--75950. Curran Associates, Inc., 2023.
\newblock URL
  \url{https://proceedings.neurips.cc/paper_files/paper/2023/file/ef7653bbc4655305efb89a32362e332a-Paper-Datasets_and_Benchmarks.pdf}.

\bibitem[Van~Horn et~al.(2018)Van~Horn, Mac~Aodha, Song, Cui, Sun, Shepard,
  Adam, Perona, and Belongie]{horn2018inaturalist}
Grant Van~Horn, Oisin Mac~Aodha, Yang Song, Yin Cui, Chen Sun, Alex Shepard,
  Hartwig Adam, Pietro Perona, and Serge Belongie.
\newblock The i{N}aturalist species classification and detection dataset.
\newblock In \emph{Proceedings of the IEEE Conference on Computer Vision and
  Pattern Recognition}, pages 8769--8778, 2018.

\bibitem[Veli{\v{c}}kovi{\'{c}} et~al.(2018)Veli{\v{c}}kovi{\'{c}}, Cucurull,
  Casanova, Romero, Li{\`{o}}, and Bengio]{velickovic2018graph}
Petar Veli{\v{c}}kovi{\'{c}}, Guillem Cucurull, Arantxa Casanova, Adriana
  Romero, Pietro Li{\`{o}}, and Yoshua Bengio.
\newblock Graph attention networks.
\newblock In \emph{International Conference on Learning Representations}, 2018.

\bibitem[Vinyals et~al.(2016)Vinyals, Bengio, and Kudlur]{set2set2016}
Oriol Vinyals, Samy Bengio, and Manjunath Kudlur.
\newblock Order matters: Sequence to sequence for sets.
\newblock In \emph{International Conference on Learning Representations}, 2016.

\bibitem[Vivanco~Cepeda et~al.(2023)Vivanco~Cepeda, Nayak, and
  Shah]{geoclip2023}
Vicente Vivanco~Cepeda, Gaurav~Kumar Nayak, and Mubarak Shah.
\newblock Geoclip: Clip-inspired alignment between locations and images for
  effective worldwide geo-localization.
\newblock \emph{Advances in Neural Information Processing Systems},
  36:\penalty0 8690--8701, 2023.

\bibitem[Wang et~al.(2025{\natexlab{a}})Wang, Cheng, Law, Zeng, Yin, and
  Liu]{wang2025caliper}
Xinglei Wang, Tao Cheng, Stephen Law, Zichao Zeng, Lu~Yin, and Junyuan Liu.
\newblock Multi-modal contrastive learning of urban space representations from
  {POI} data.
\newblock \emph{Computers, Environment and Urban Systems}, 118:\penalty0
  102299, 2025{\natexlab{a}}.
\newblock \doi{10.1016/j.compenvurbsys.2025.102299}.

\bibitem[Wang et~al.(2025{\natexlab{b}})Wang, Xiong, Liu, Stewart, Dujardin,
  Bountos, Zavras, Gerken, Papoutsis, Leal-Taixé, and
  Zhu]{wang2025unifiedcopernicusfoundationmodel}
Yi~Wang, Zhitong Xiong, Chenying Liu, Adam~J. Stewart, Thomas Dujardin,
  Nikolaos~Ioannis Bountos, Angelos Zavras, Franziska Gerken, Ioannis
  Papoutsis, Laura Leal-Taixé, and Xiao~Xiang Zhu.
\newblock Towards a unified copernicus foundation model for earth vision,
  2025{\natexlab{b}}.
\newblock URL \url{https://arxiv.org/abs/2503.11849}.

\bibitem[Wen et~al.(2025)Wen, Cai, Ma, Li, Chen, Webster, and Zhou]{mora2025}
Ya~Wen, Jixuan Cai, Qiyao Ma, Linyan Li, Xinhua Chen, Chris Webster, and Yulun
  Zhou.
\newblock {MoRA}: Mobility as the backbone for geospatial representation
  learning at scale.
\newblock \emph{arXiv preprint arXiv:2506.01297}, 2025.

\bibitem[Wo{\'z}niak and Szyma{\'n}ski(2021)]{wozniak2021hex2vec}
Szymon Wo{\'z}niak and Piotr Szyma{\'n}ski.
\newblock Hex2vec: Context-aware embedding {H3} hexagons with {OpenStreetMap}
  tags.
\newblock In \emph{Proceedings of the 4th ACM SIGSPATIAL International Workshop
  on AI for Geographic Knowledge Discovery}, pages 61--71, 2021.

\bibitem[Yan et~al.(2024)Yan, Wen, Zhong, Chen, Chen, Wen, Zimmermann, and
  Liang]{urbanclip}
Yibo Yan, Haomin Wen, Siru Zhong, Wei Chen, Haodong Chen, Qingsong Wen, Roger
  Zimmermann, and Yuxuan Liang.
\newblock Urbanclip: Learning text-enhanced urban region profiling with
  contrastive language-image pretraining from the web.
\newblock In \emph{Proceedings of the ACM Web Conference 2024}, WWW '24, page
  4006–4017, New York, NY, USA, 2024. Association for Computing Machinery.
\newblock ISBN 9798400701719.
\newblock \doi{10.1145/3589334.3645378}.
\newblock URL \url{https://doi.org/10.1145/3589334.3645378}.

\end{thebibliography}
\end{document}